\documentclass[runningheads]{llncs}
\usepackage{graphicx}
\usepackage{adjustbox}
\usepackage{amsmath,amssymb} 
\usepackage{subfigure}
\usepackage{color}
\usepackage[width=122mm,left=12mm,paperwidth=146mm,height=193mm,top=12mm,paperheight=217mm]{geometry}
\begin{document}
\pagestyle{headings}
\mainmatter

\title{Towards Automatic Wild Animal Monitoring: Identification of Animal Species in Camera-trap Images using  Very Deep Convolutional Neural Networks} 

\titlerunning{.}

\authorrunning{A. Gomez, A. Salazar and F. Vargas}

\author{Alexander G\'omez\inst{1} \and Augusto Salazar\inst{1,}\inst{2} \and Francisco Vargas\inst{1}}

\institute{Grupo de investigaci\'on SISTEMIC, Facultad de Ingenier\'ia, Universidad de Antioquia UdeA, Calle 70 No. 52 -- 21, Medell\'in, Colombia.\and Grupo de investigaci\'on AEyCC, Facultad de Ingenier\'ias, Instituto Tecnol\'ogico Metropolitano ITM, Carrera 21 No. 54-10, Medell\'in, Colombia. \\
\email{alexander.gomezv@udea.edu.co }}

\maketitle

\begin{abstract}

Non intrusive monitoring of animals in the wild is possible using camera trapping framework, which uses cameras triggered by sensors to take a burst of images of animals in their habitat. However camera trapping framework produces  a high volume of data (in the order on thousands or millions of images), which must be analyzed by a human expert. In this work, a method for animal species identification in the wild using very deep convolutional neural networks is presented. Multiple versions of the Snapshot Serengeti dataset were used in order to probe the ability of the method to cope with different challenges that camera-trap images demand. The method reached $88.9\%$ of accuracy in Top-1 and $98.1\%$ in Top-5 in the evaluation set using a residual network topology. Also, the results show that the proposed method outperforms previous approximations and proves that recognition in camera-trap images can be automated.

\end{abstract}


\section{Introduction}
\label{sec:Introduction}

Since its invention in the Nineteenth Century photography has been used for multiple science purposes including astronomy, medicine and biology. To observe wild animals without disturbing them was an early goal of photography, but it was not until 1890 when George Shiras developed a method using a tripwire and a flash system allowing that wild animals be auto-photographed~\cite{shiras1906photographing}. Shiras' method was the beginning of camera trapping development that evolved to more complex systems and that nowadays uses infrared beams as a triggering device. They are small, portable, digital, and  resistant to the most challenging ecosystem conditions for weeks, yet months.

Currently, automated camera-traps used in wildlife studies are small boxes, grabbed to a tree, rock or other structure. Camera-traps are powerful tools for wildlife scientists who, thanks to this method, can answer questions such as ``Which animal species occurs in a certain area?'', ``What are they doing?'', ``How many are there?'', among others. Also fundamental studies, like detecting rare species, delineating species’ distributions, documenting predation, monitoring animal behavior, and other vital rates~\cite{o2010camera} are carried out with this method. Hence, it allows biologist to protect animals and its environment from extinction or man-made damage.

Although camera trapping is a useful methodology in ecology this method generates a large volume of images. Therefore it is a big challenge to process  the recorded images and even harder, if the biologists are looking to identify all photographed species. Currently, no automatic approach is used to identify species from camera-trap images. Researchers analyze  thousands or millions of photographs manually~\cite{fegraus2011data}. An automatic system that deals with this problem would accelerate the professionals' work, allowing them to focus on data analysis and important issues only. 

Automatic classification of animal species in camera-trap images still remains  an unsolved problem due to very challenging image conditions. A few previous works proposed solutions for this problem. Yu et.al~\cite{yu2013automated} manually cropped and selected images, which contain the whole animal body. This conditioning allowed then to obtain $82\%$ of accuracy classifying $18$ animal species in their own dataset. Chen et.al~\cite{chen2014deep} use an automatic segmentation algorithm but  they obtained only $38.3\%$ of accuracy. Before the releasing of the Snapshot Serengeti dataset~\cite{swanson2015snapshot} there was no publicly available dataset to work with and for benchmarking.  This dataset  was released in 2015 to allow the computer science community to study and overcome the challenges present in  camera trapping  framework. In this work very deep convolutional neural networks are proposed in order to solve the species identification task over the Snapshot Serengeti dataset. Versions of unbalanced, balanced, conditioned,  and segmented dataset are used in order to study if a powerful learning algorithm can overcome the four common issues in camera trapping: unbalanced samples, empty frames, incomplete animal images, and objects too far from focal distance. Our results show that with enough data, conditions  and learning capacity, the camera trapping species recognition task can be fully automated. The model also outperforms previous approximations  and shows how robust learning algorithms are to corrupted data produced by wrong annotations from citizens. In addition all datasets derived from Snapshot Serengeti (including manually segmented images) and the trained models, are publicly available.

The  rest of the paper is organized as follows: First, related work is mentioned in section~\ref{sec:Related_Work}. In section~\ref{sec:Methods} the challenges present in camera trapping framework  are described. Also the methods used in the identification model are explained. Section 4 describes the experiments used to test the models. Results are presented in section~\ref{sec:Results}. Section \ref{sec:discussion} discusses the results. Finally, in section~\ref{sec:Conclusions} conclusions and future work are presented.

\section{Related Work}
\label{sec:Related_Work}

This section reviews previous approaches to identify multiple species in camera-trap images. To the best of our knowledge there are only two previous approaches to identify animal species in camera-trap images. Sparse coding spatial pyramid matching (ScSPM) was used by Yu et.al~\cite{yu2013automated} to recognize $18$ species of animals, reaching $82\%$ of accuracy on  their own dataset (composed of $7196$ images). The ScSPM  extract dense SIFT descriptors and cell-structured local binary patterns as local features; then global features are generated using global weighted sparse coding and max pooling thought multi-scale pyramid kernel. The images are classified  using a linear support vector machine. As input to the ScSPM  photo-trap images were preprocessed: Removing empty frames (images without animals), manually cropping all the animals from the images, and selecting only those images that captures the animals' whole body.

A deep convolutional neural network (ConvNet) was used by Chen et.al~\cite{chen2014deep} to classify 20 animal species in their own dataset. An important difference from ~\cite{yu2013automated} is that they use an automatic segmentation method (Ensemble Video Object Cut) for cropping the animals from the images and use this crops to train and test their system. The ConvNet used only has $6$ layers ($3$ convolutional layers and $3$ max pooling layers) which give them a $38.31\%$ of accuracy.

Our approach uses ConvNets as~\cite{chen2014deep} but is different in two main aspects. First, an analysis using an unbalanced, balanced, conditioned, and segmented dataset is done . Very deep ConvNets (AlexNet~\cite{krizhevsky2012imagenet}, VGGNet~\cite{simonyan2014very}, GoogLenet~\cite{szegedy2015going} and ResNets~\cite{He2015}) were used in order to probe that with a higher learning capacity and enough images, the camera trapping recognition problem can be fully automated. In addition, our manually segmented images differ from the work of Yu et.al~\cite{yu2013automated} in the sense that their crops contain the whole animal body. In contrast, our crops  also contain images containing only some body parts, which, as will be explained later, add high complexity to the classification task.


\section{Towards  animal monitoring in the wild }
\label{sec:Methods}

In this section different situations present in camera-trap images that must be overcome to make species identification automatically, are described and analysed. Also, a solution based on very deep convolutional neural networks is proposed.

\subsection{Challenges in camera trapping}

\begin{figure}[hbtp]
\centering			
\subfigure[]{\includegraphics[width=0.19\textwidth]{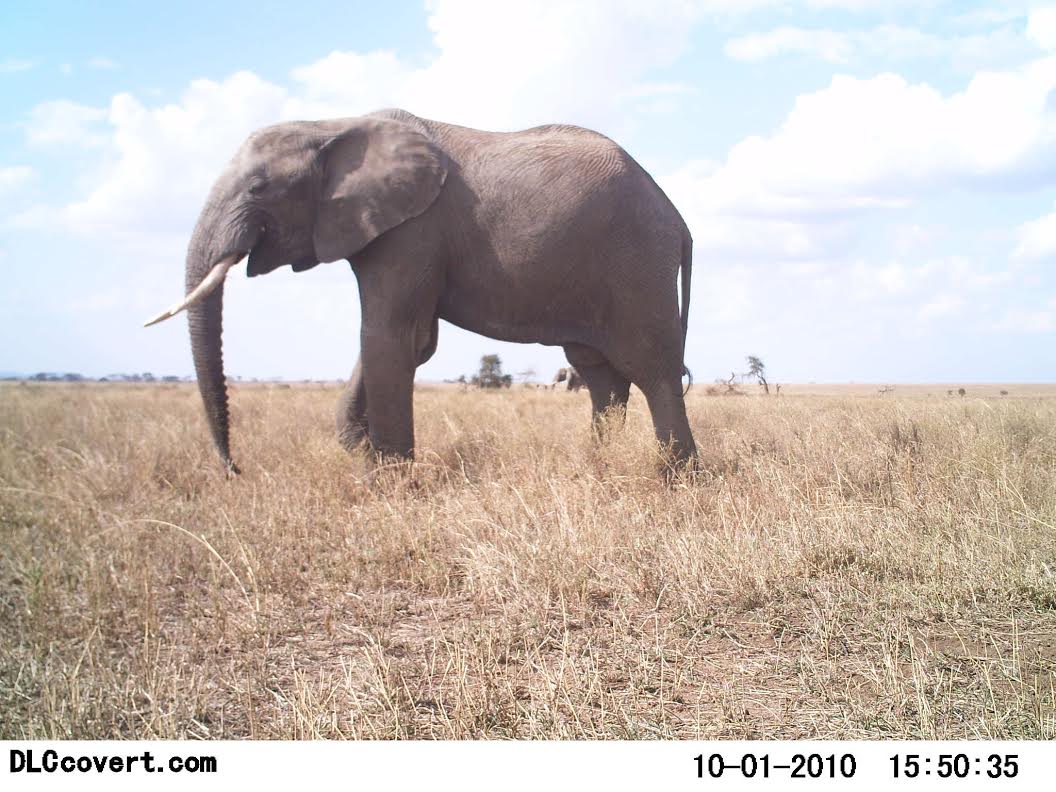}{\label{fig:preprocessing1}}}
\subfigure[]{\includegraphics[width=0.19\textwidth]{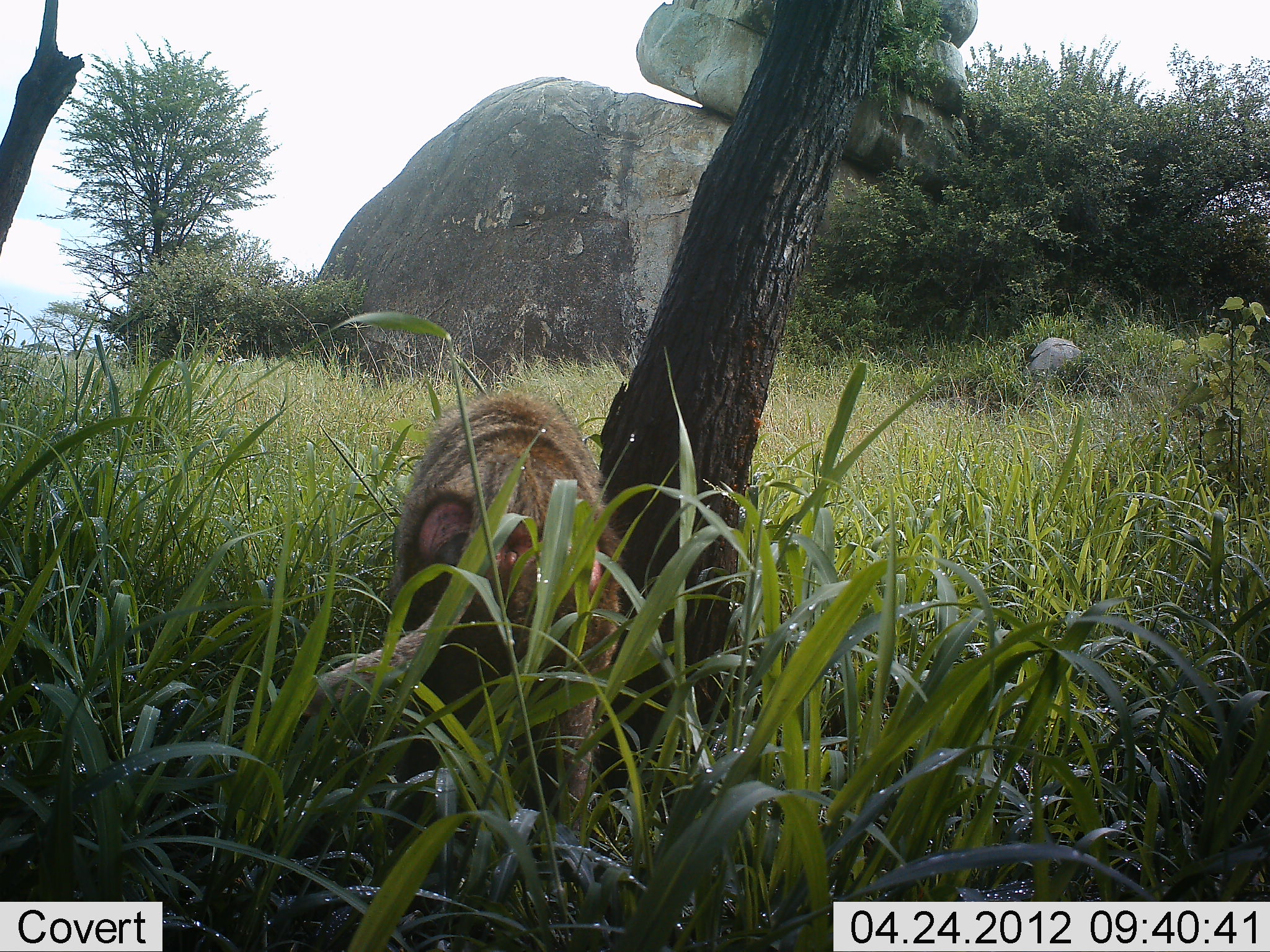}{\label{fig:preprocessing2}}}
\subfigure[]{\includegraphics[width=0.19\textwidth]{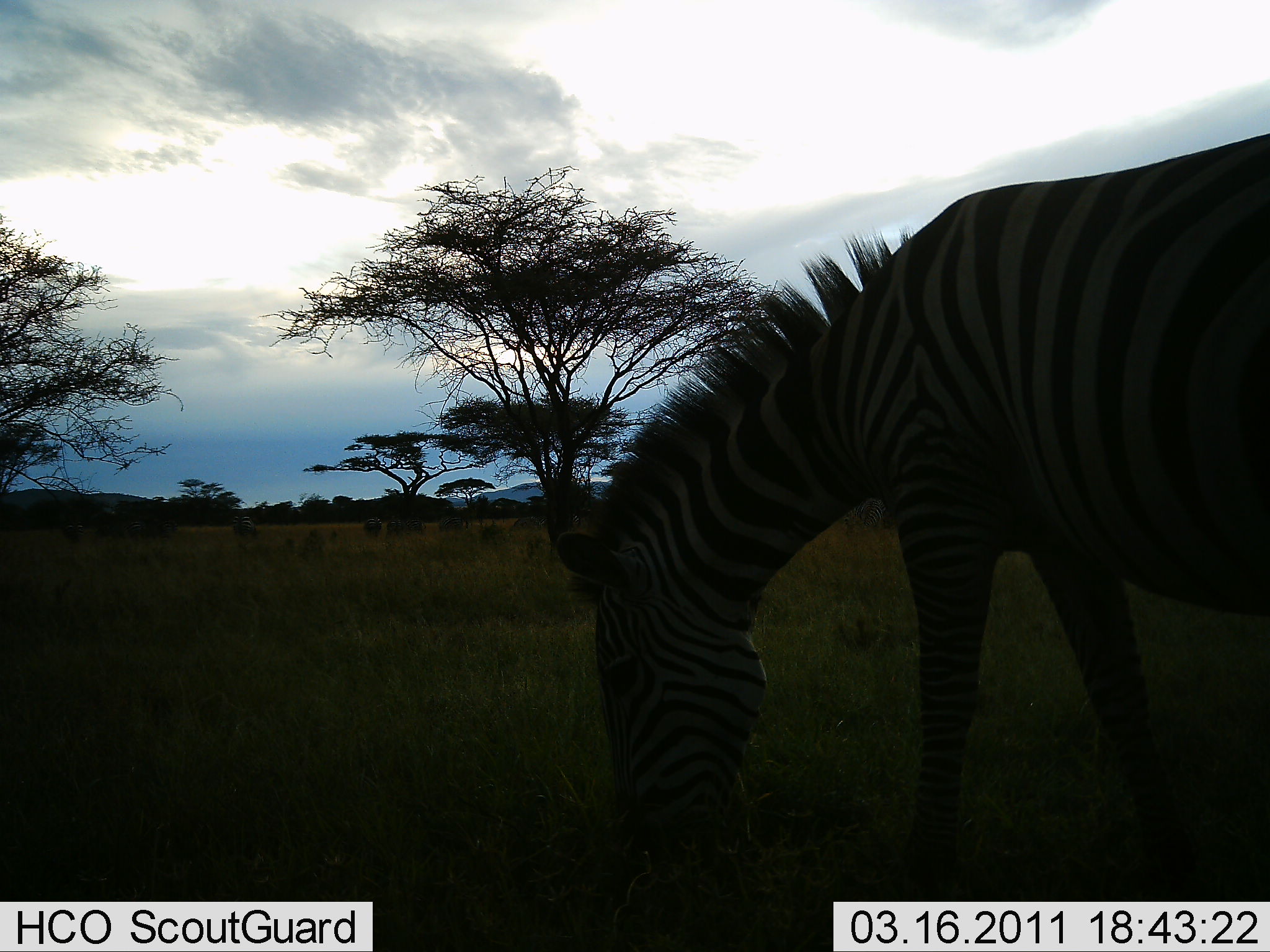}{\label{fig:preprocessing3}}}
\subfigure[]{\includegraphics[width=0.19\textwidth]{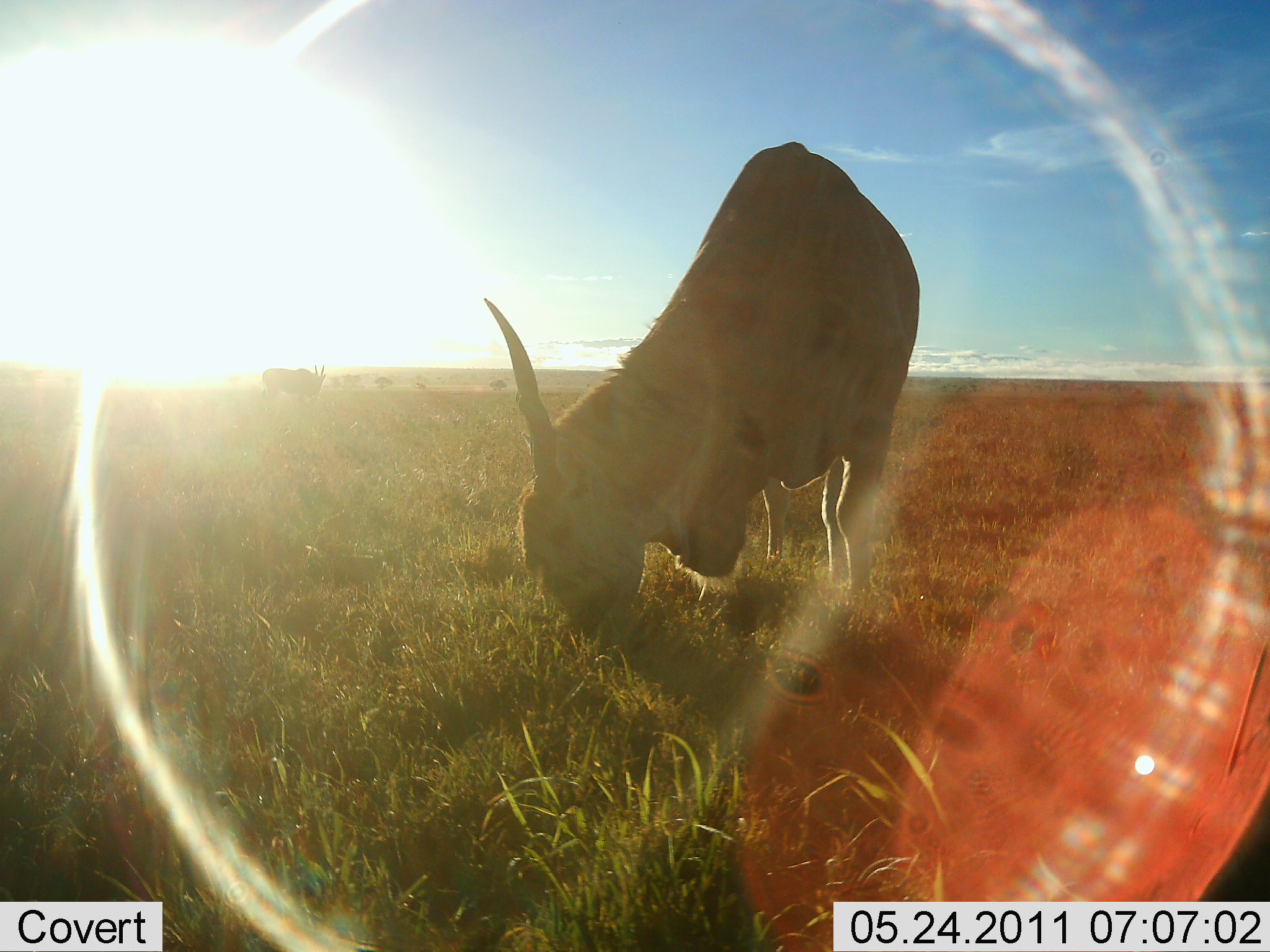}{\label{fig:preprocessing4}}}
\subfigure[]{\includegraphics[width=0.19\textwidth]{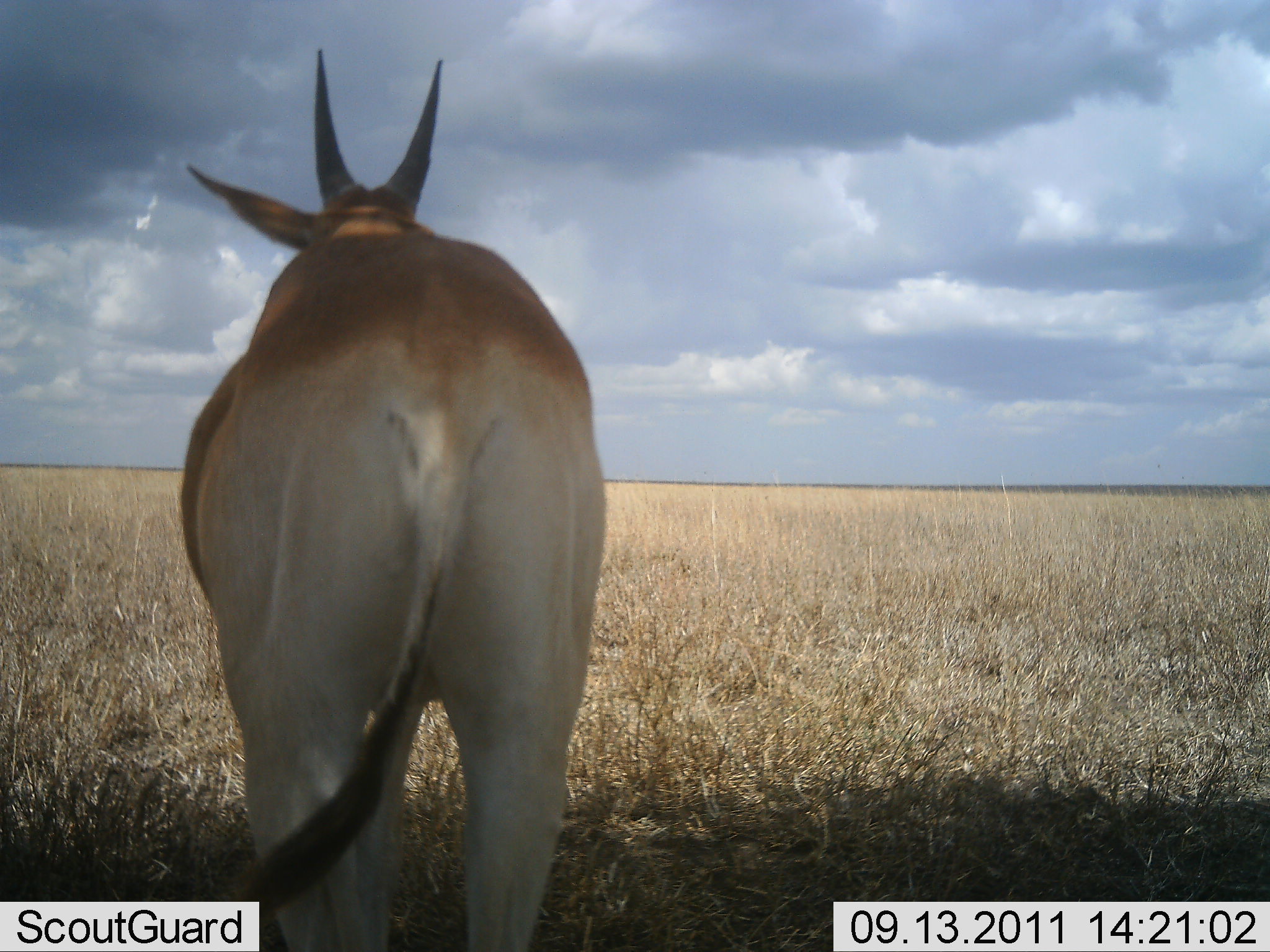}{\label{fig:preprocessing5}}}
\subfigure[]{\includegraphics[width=0.19\textwidth]{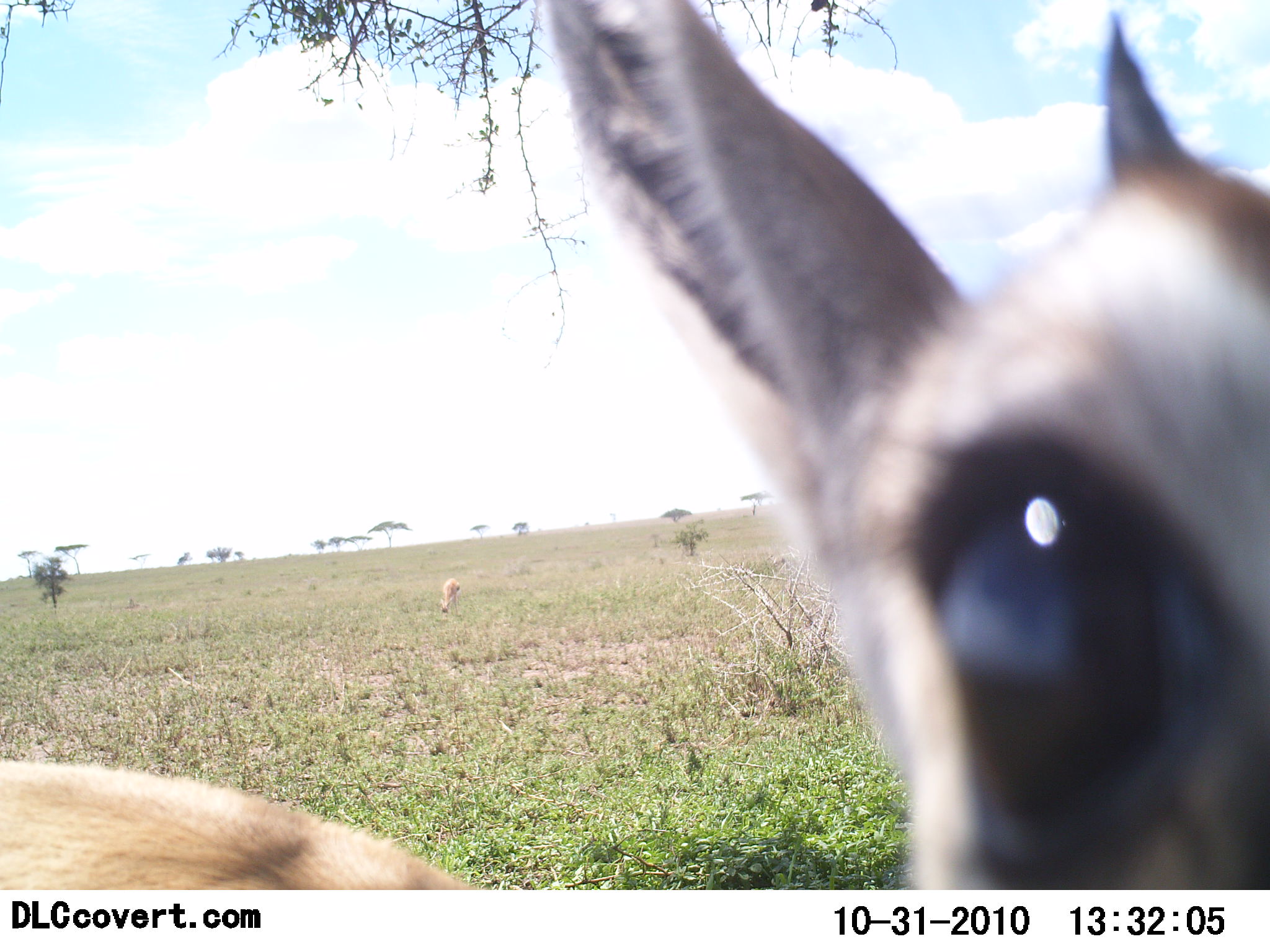}{\label{fig:preprocessing6}}}
\subfigure[]{\includegraphics[width=0.19\textwidth]{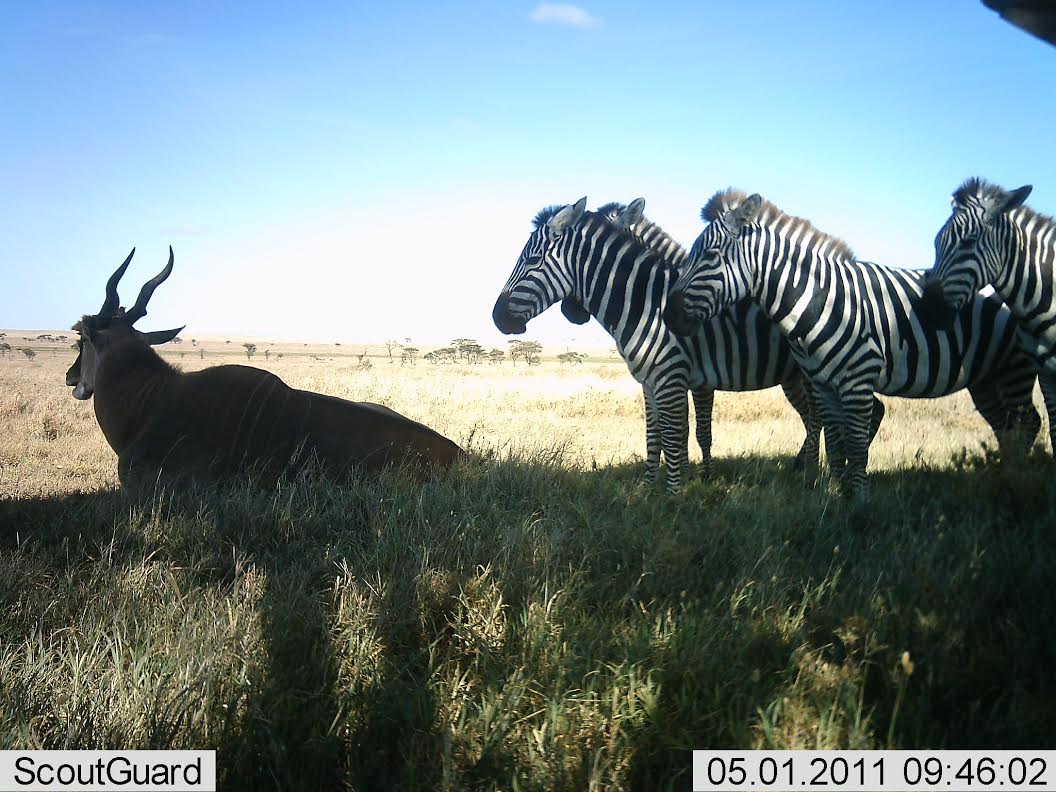}{\label{fig:preprocessing7}}}
\subfigure[]{\includegraphics[width=0.19\textwidth]{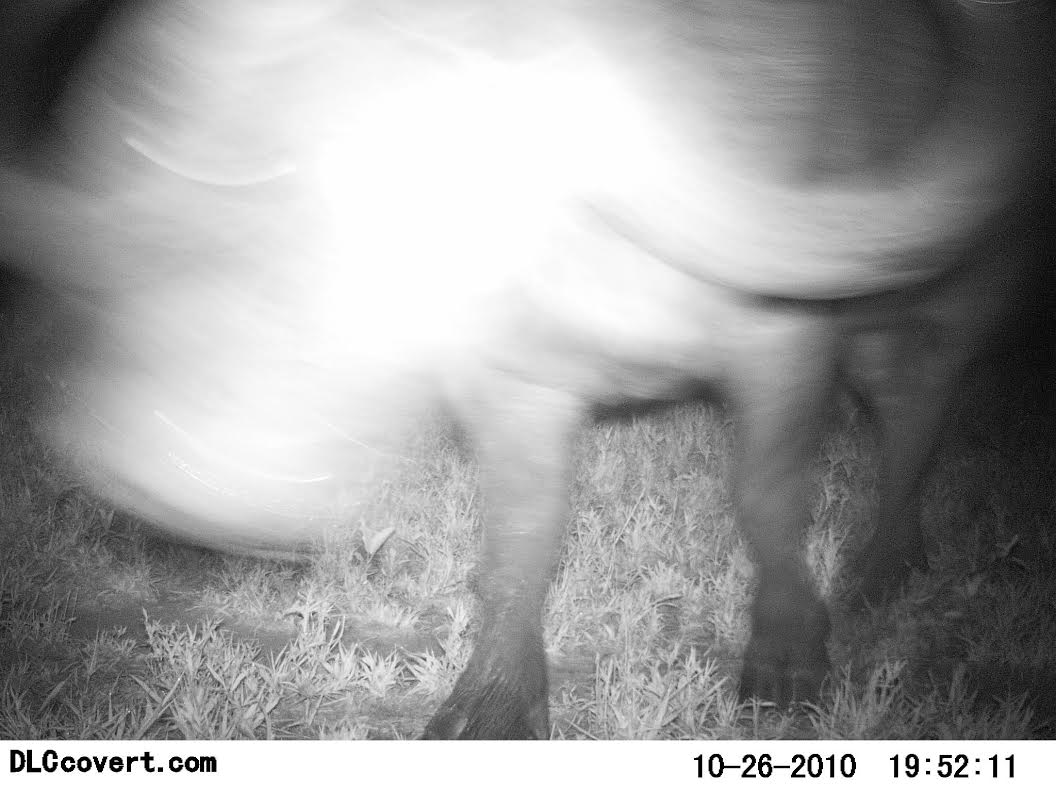}{\label{fig:preprocessing8}}}
\subfigure[]{\includegraphics[width=0.19\textwidth]{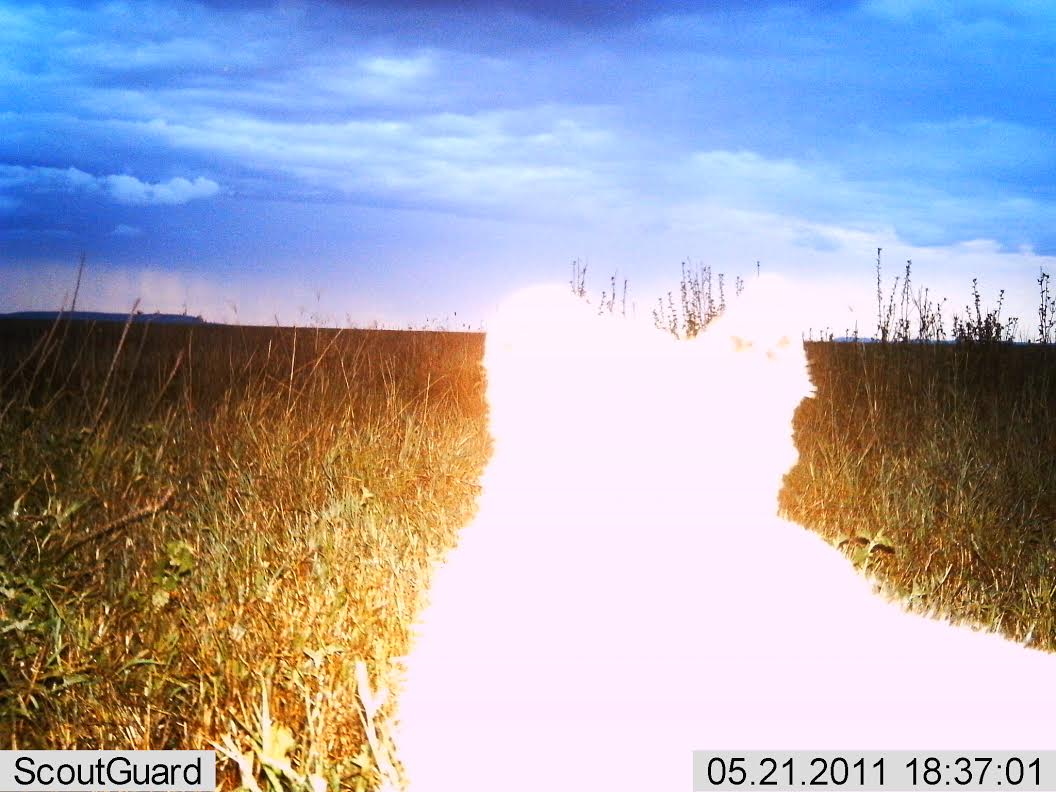}{\label{fig:preprocessing9}}}
\subfigure[]{\includegraphics[width=0.19\textwidth]{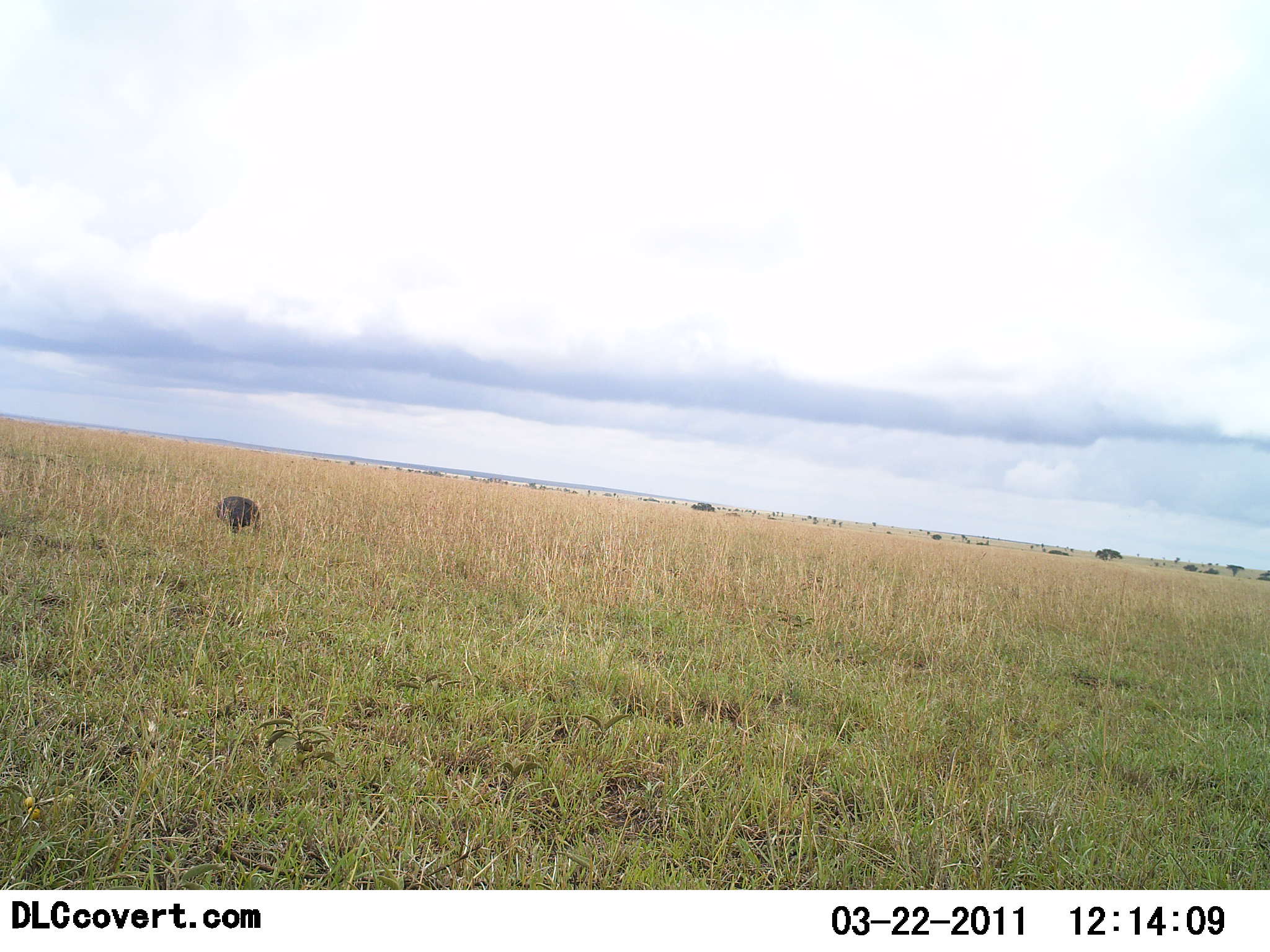}{\label{fig:preprocessing10}}}
\caption{Different camera trapping classification scenarios. (a) Ideal. (b) Occlusion due to context. (c) Poor illumination. (d) Over-exposed regions. (e) Auto-occlusion. (f) Complex animal poses and unexpected images. (g) Different species in the same image. (h) Blurred. (i) Over-exposed animals. (j) Animals far way from camera. All images were take from Snapshot Serengeti dataset\cite{swanson2015snapshot} as well as all photo-trap images used in this document.
}
\label{fig:preprocessing}
\end{figure}

Recognition of animal species in camera-trap images can be interpreted as a object recognition problem. In this case, for instance an elephant is present in the image and it must be localized and classified as elephant (see Fig.~\ref{fig:preprocessing1}). Notice that images like the elephant are an ideal and scarce case. In camera trapping the challenges can be classified in three main groups: Environmental conditions, animal behaviour related, and hardware limitations.

Environmental conditions denotes how the context affects the quality of a camera-trap image. Since the camera-traps are set in the wild and remain there for long periods, many objects can occlude animals  as Fig.~\ref{fig:preprocessing2} shows. As environment does not remain equal (e.g., plants grow, trees fall, among others)  even if when the cameras were placed, there was no occlusion, it can appear at any moment. Day and night have different illumination conditions but the transition between them also causes problems (see Fig.~\ref{fig:preprocessing3}).
Fig.~\ref{fig:preprocessing4} shows an example of overexposed regions caused by the sun light. Variations like  rain and drops in the lens are also examples of conditions that directly affect hardware performance.

A common approach in camera trapping framework is to put the camera pointing to a natural path or to an expected place where the animals will pass through. However, animals do not behave in a predictable way. A proof of this statement is the very few images that contains animals in an ideal condition. A vast majority of camera-trap images do not contain the whole body of an animal due to: Context occlusion, the animal is too close to the camera (see Fig.~\ref{fig:preprocessing6}), among others. Also animals are captured in random poses as Fig.~\ref{fig:preprocessing5}  and Fig.~\ref{fig:preprocessing6} show, which hide important features to recognize an species, and reduce confidence in a classification decision. Particularly in similar species this fact is crucial as we will discuss later. Finally, it is also possible that several species appear in one single image (see Fig.~\ref{fig:preprocessing7}). This adds high complexity to the recognition problems, because it forces to localize all animals in the image. The above mentioned situations could happen independently or simultaneously in all  possible combinations.

Camera trapping hardware assembles high resistant and low power consumption electronics. There are several models commercially available, which allow different parameter configurations. Resolution of the camera, trigger sensor, frames per second, and night illumination (infra-red or flash) are some of the main factors when selecting a camera. Blurred images and overexposed animal images,  as Fig.~\ref{fig:preprocessing8} and  Fig.~\ref{fig:preprocessing9} show, depends on camera-trap hardware and selected parameters (time between frames and flash power for instance). These  problems  make shape and fur patterns indistinguishable  increasing  the identification task difficulty. 

\begin{figure}[hbtp]
\centering			
\subfigure[]{\includegraphics[width=0.24\textwidth]{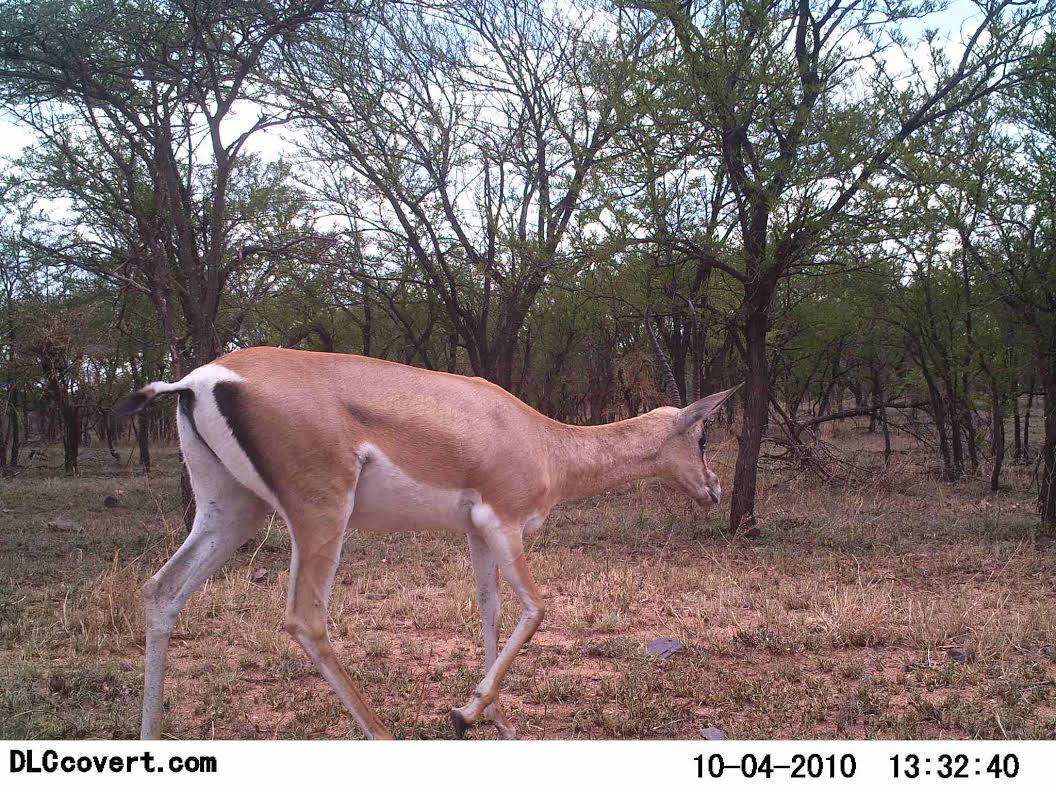}{\label{fig:gazelleG}}}
\subfigure[]{\includegraphics[width=0.24\textwidth]{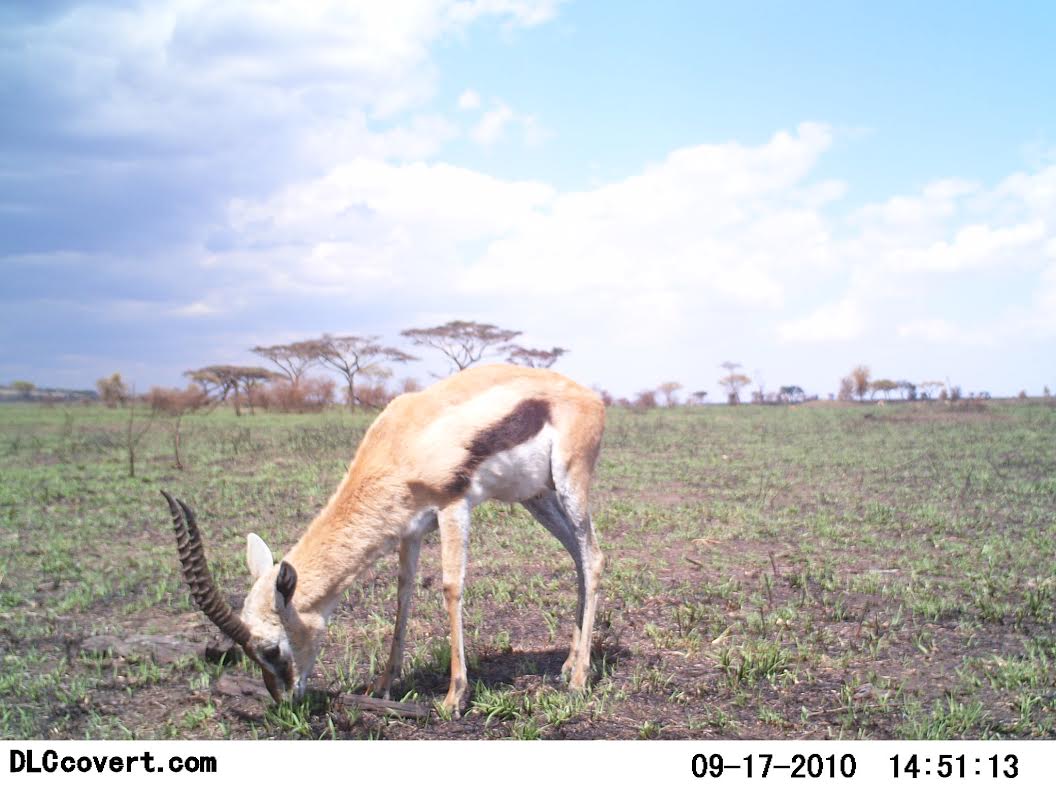}{\label{fig:GazelleT}}}
\subfigure[]{\includegraphics[width=0.24\textwidth]{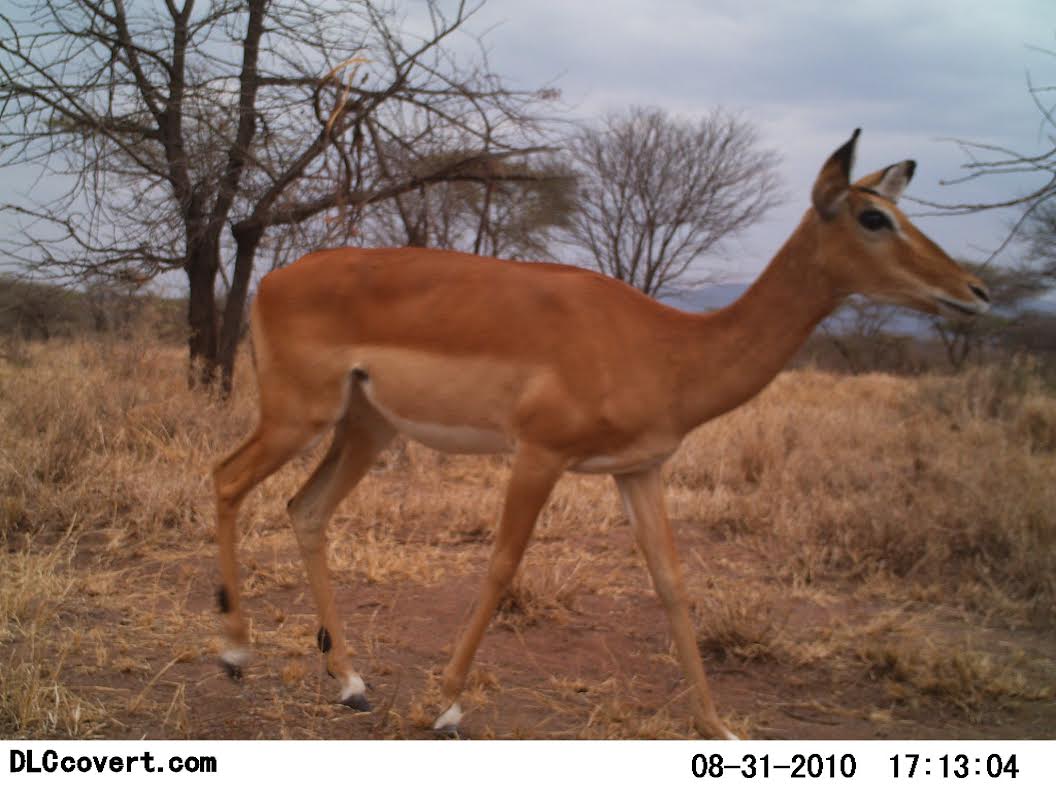}{\label{fig:impala}}}
\subfigure[]{\includegraphics[width=0.24\textwidth]{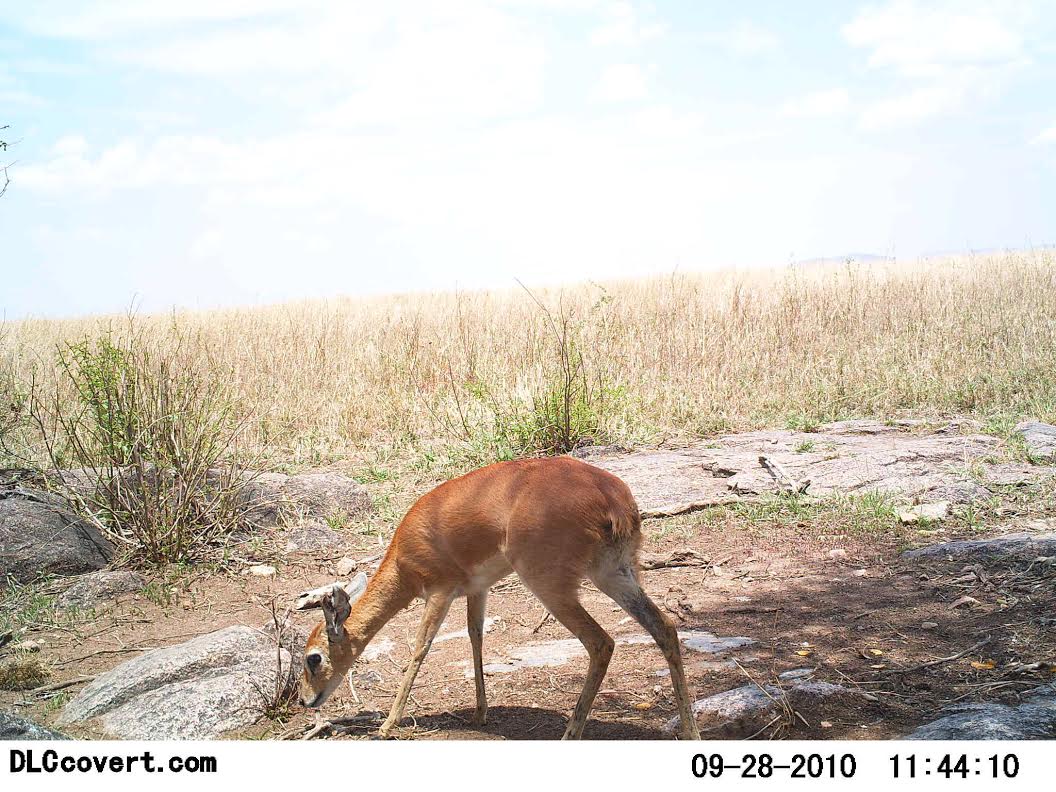}{\label{fig:reedbuck}}}
\subfigure[]{\includegraphics[width=0.24\textwidth]{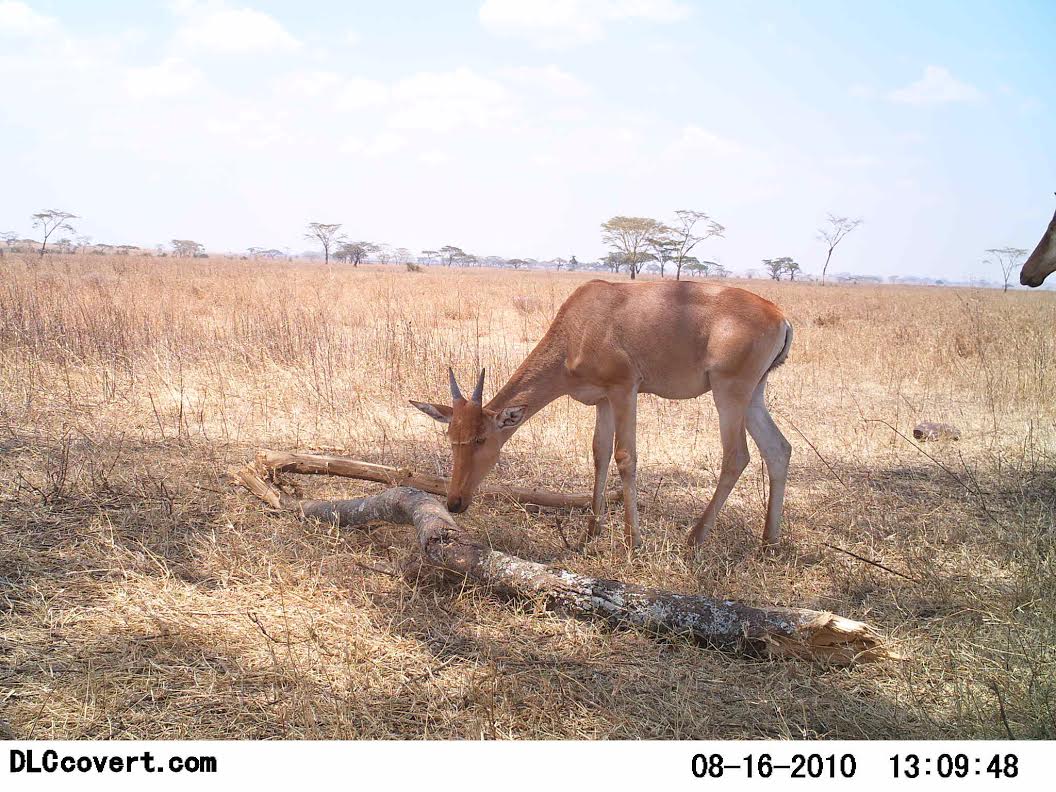}{\label{fig:buffalo}}}
\subfigure[]{\includegraphics[width=0.24\textwidth]{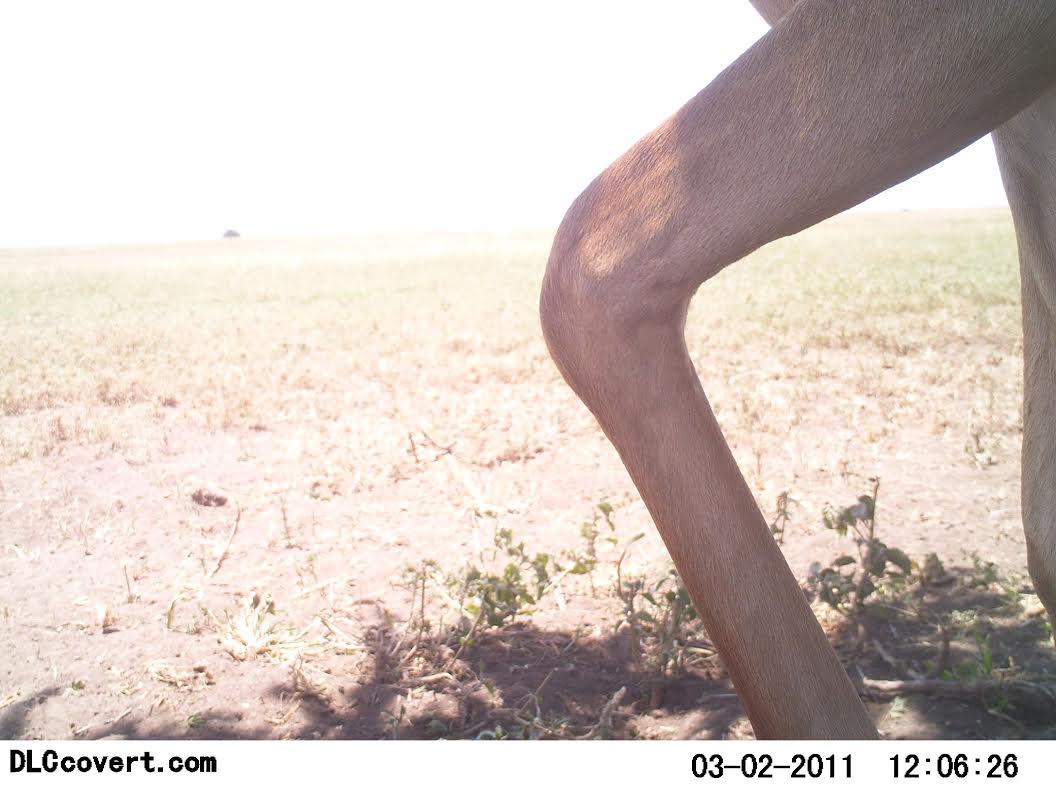}{\label{fig:impalaLeg}}}
\subfigure[]{\includegraphics[width=0.24\textwidth]{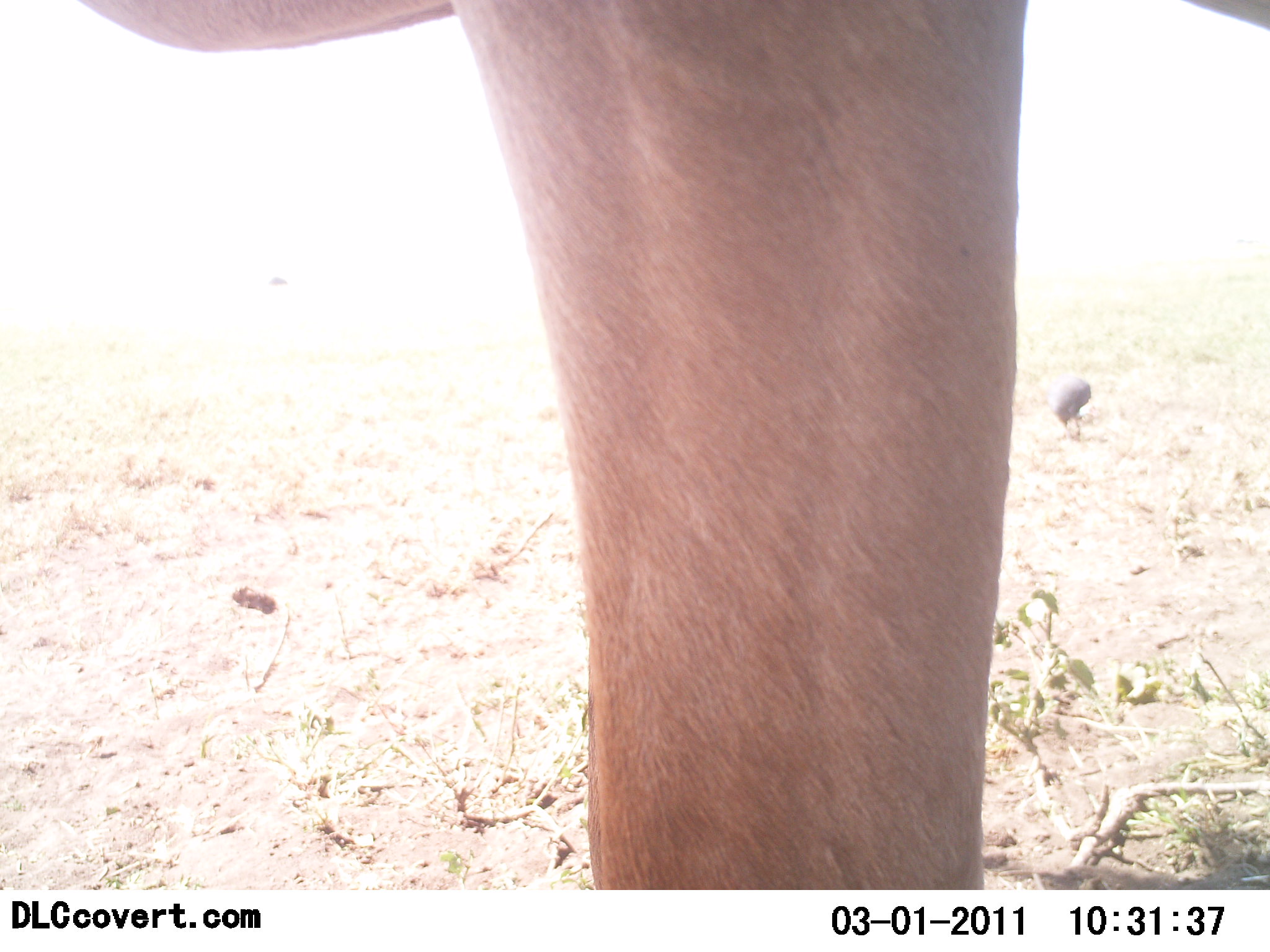}{\label{fig:impalaLeg2}}}
\subfigure[]{\includegraphics[width=0.24\textwidth]{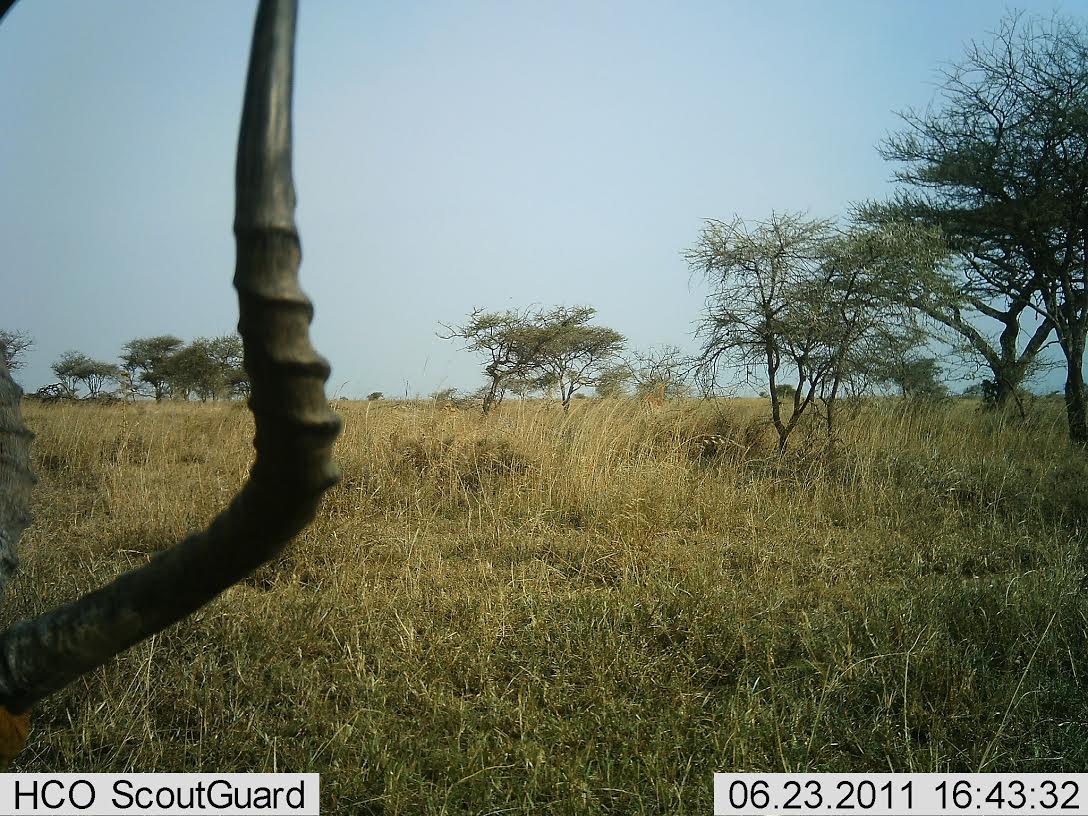}{\label{fig:impalaLeg3}}}
\caption{(a) Gazelle Grant. (b) Gazelle Thompson. (c) Impala. (d) Reedbuck. (e)  Eland  (f) Impala's leg. (g) Impala's leg. (f) Impala's horn. 
}
\label{fig:preprocessing}
\end{figure}

Despite image condition, resembling species recognition turn on the problem in a fine-grained classification task. In  Fig.~\ref{fig:gazelleG} to Fig.~\ref{fig:buffalo} five similar species found in the same area are show. Partial (mainly because of inter-frame time or animals close to the camera) and not expected poses can make fine grained classification harder (see Fig.~\ref{fig:impalaLeg} and Fig.~\ref{fig:impalaLeg2}).   

An ideal automatic species recognition system must deal with all above mentioned conditions. Also it  must classify an animal only with partial information as Fig.~\ref{fig:impalaLeg3} shows. Very deep convolutional neural networks are the state of art in image recognition and are known for their high learning capacity (able to learn up to 1000 classes~\cite{He2015}) and robustness to typical object recognition challenges. The next section describes the use of convolutional neural networks to solve the species recognition problem.

\subsection{Convolutional Neural Networks}

Convolutional neural networks~\cite{lecun1998gradient} consist of stacked convolutional and pooling layers ending in a  fully conected layer with  a feature vector as Fig.3 shows as output. Convolutional layers generate feature maps followed by a non-linear activation function.  Pooling layers provides scale invariant capacity to the extracted features. A common topology in a ConvNet consists of many sequential stacked convolutional and pooling layers that can extract discriminative features from an input image.

\begin{figure}[htpb]
\centering
\label{fig:conv}
\includegraphics[width=\textwidth]{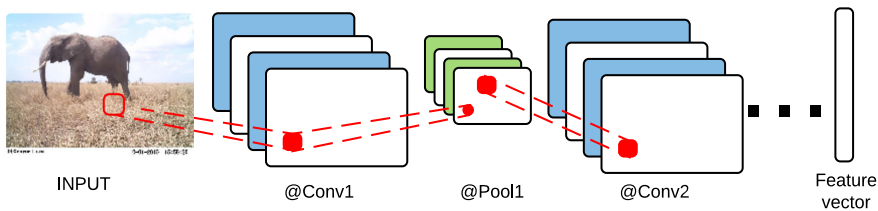}
\caption{Convolutional Neural Network}
\end{figure}

A transformation that maps from low level to high level features is done in a ConvNet.  The first layers contain low level features (e.g., edges and orientation) and the last layers contain high level representation features, such as wrinkles, or in the case of animals, the fur details and its discriminative patterns. An important issue in ConvNets architectures is the Depth, reason why community attempts to boost topology Depth. In this work AlexNet~\cite{krizhevsky2012imagenet}, VGGNet~\cite{simonyan2014very}, GoogLenet~\cite{szegedy2015going}, and ResNets~\cite{He2015} are used in order to probe how Depht in ConvNets impacts in camera trapping species recognition.

AlexNet is a well known model of ConvNet due to its success in the ImageNet classification challenge~\cite{krizhevsky2012imagenet}. It consists of $5$ convolutional layers, $3$ fully connected layers, and a dropout step in training to deal with overfitting. In the search for more discriminative models, the VGGNet was created. It reached $16$ and $19$ layers of depth  using very small convolutional filters in all convolutional layers. With the same purpose, GoogLenet, in addition to the convolutional and pooling layers, includes the inception module. This new module extracts non-linear feature maps and adds a sparse connection concept to deal with network size and allows GoogLenet to reach $22$ layers. Finally, residual networks address the vanishing gradients problem (a common issue when the Depth increases in deep ConvNets~\cite{glorot2010understanding}) introducing a deep residual learning framework that adds short-cut connections between layers allowing ConvNet to have until $1000$ layers of depth.

In this work ConvNets are used as black-box feature extractors (using off-the-shelf features~\cite{razavian2014cnn}). Also the ConvNets are fine-tuned to fit  our classification problem. The black-box feature extractor consists of the use of features from highest layers of a ConvNet as inputs of a linear classifier without modifying or training the ConvNet again. This approach is especially useful when the ConvNet  was trained with images similar to the new classification problem. A trained ConvNet can be fine-tuned which means  to run the back-propagation algorithm in the pre-trained version of the Net. Its expected that the previously knowledge helps to  improve the learning of the new classification problem. Black box feature extractor and fine-tune- process involve a concept called transfer learning~\cite{pan2010survey}.

\section{Experimental Framework}
\label{sec:Experimental_Framework}

In this section both, the datasets used and the experiments carried out in this work, are described. Additionally, an explanation of implementation details (such as libraries and architecture parameters) is included.

\subsection{Datasets}

The Snapshot Serengeti dataset~\cite{swanson2015snapshot} is a camera-trap database published in 2015. It was acquired with 225 camera-traps placed in Serengeti National Park, Tanzania. There were taken more than one million sets of pictures, each set containing $1-3$ photographs taken in a single burst of one second. A particularity of Snapshot Serengeti is that persons from general public annotated the images from a set called Consensus and experts annotate a sub-set of consensus dataset called Gold Standard.

The Consensus set was validated against Gold Standard revealing $96.6\%$ accuracy for species identification. This is  a successful example of citizen science application. In this work the consensus set is used to train and validate the models.

The Snapshot Serengeti contains images of 48 animal species. It is a highly unbalanced dataset, e.g., zebra class has $179683$ images and the striped polecat (zorilla) only $29$. In this work only $26$ classes were selected for classification (they are listed in Table~\ref{tab:animals}). Each of the selected classes has enough images to train the models in a balanced way. 
 
In order to compare with previous approaches, the camera-trap dataset used in~\cite{chen2014deep} was also used (this dataset will be referenced as Panama dataset in the following). It contains $20$ species common to North America and has $14346$ training images and $9530$ testing images. These images were generated using an automatic segmentation algorithm described in~\cite{ren2013ensemble}. The segmentation algorithm uses the burst of camera-trap photos to segment the moving objects. The Panama dataset contains color and infrared images. A particularity of this dataset (or at least the one that they shared with the authors of this paper) is that it also contains images without animals that have a species label due to the fact that the segmentation algorithm does not work perfectly. No photos were eliminated or selected.

\subsection{Experiments}

The $26$ selected classes from the Consensus set were divided into several versions (see Table~\ref{tab:dataset}). Tests using $D1$ dataset are aimed to know how well the models deal with the highly unbalanced nature of camera trapping databases. This dataset is composed of all images listed in Table~\ref{tab:animals}. To evidence the impact of unbalanced data, the $D2$ dataset is used.  The goal of this experiment is to see if the models are able to fit easier than in the unbalanced case. This is a subset from $D1$, where $1000$ and $240$ images per class were selected as train and evaluation sets, respectively. Dataset $D3$ was generated from images where the animal is placed in the foreground. This conditions are ideal and difficult to reach in practice. For instance, a sensor with a shorter range could be used, but this is not a wanted condition for biologists since they want to capture as many animal photos as possible. Finally, dataset $D4$ consists of manually segmented images to simulate a segmentation algorithm that always finds the animal (or part of it) in the image. This conditioning  differs from  Yu et.al's~\cite{yu2013automated} cropped images, where the cropped images only contain whole bodies of animals. In our case the cropped images include images where the animals appear partially. 

\begin{table}[]
\centering
\caption{Species selected from Snapshot Serengeti dataset}
\label{tab:animals}

\begin{tabular}{lc@{\hskip 0.5in}lc}
\hline
\multicolumn{1}{c}{Specie}          & \#  images & \multicolumn{1}{c}{Specie}           & \# images \\
\hline
Baboon          & 4618       & Jackal            & 1207      \\
Buffalo         & 34684      & Kori bustard      & 2042      \\
Cheetah         & 3354       & Lion female\&cub  & 8773      \\
Dik-dik         & 3364       & Lion male         & 2413      \\
Eland           & 7395       & Ostrich           & 1945      \\
Elephant        & 25294      & Reedbuck          & 4131      \\
Giraffe         & 22439      & Secretary bird    & 1302      \\
Grant’s gazelle & 21340      & Spotted hyena     & 10242     \\
Guinea fowl     & 23023      & Thomson's gazelle & 116421    \\
Hartebeest      & 15401      & Topi              & 6247      \\
Hippopotamus    & 3231       & Warthog           & 22041     \\
Human           & 26557      & Wildebeest        & 212973    \\
Impala          & 22281      & Zebra             & 181043    \\
\hline
\end{tabular}
\end{table}

\begin{table}[h]
\centering
\caption{Snapshot Serengeti dataset partitions}
\label{tab:dataset}
\begin{tabular}{cl}
\hline
\textbf{ Label} &  \multicolumn{1}{c}{\textbf{Description}}                       \\
\hline
$D1$                       & 26 Classes unbalanced                      \\
$D2$                       & 26 Classes balanced                        \\
$D3$                       & 26 classes objects in foreground \\
$D4$                       & 26 classes objects animals segmented \\
\hline
\end{tabular}
\end{table}

Table \ref{tab:convnets} shows the six very deep ConvNets  used in this work. They are the state of the art in object recognition. Fine-tuning was not always possible due to our hardware limitations but the results will show how fine-tuning impacts on the system performance. This work uses multiple very deep ConvNets  in order to probe how  the Depth in ConvNets impacts on the camera trapping classification problem and to answer questions such as ``Could a deeper model deal with empty frames?'', ``Could a deeper model deal with an unbalanced dataset?'', among others. In this sense for each dataset one experiment was carried out with each of the architectures listed in Table~\ref{tab:convnets} giving a total of $32$ experiments.

To compare with previous approaches an experiment C1 was done. In this experiment the model with the highest accuracy using the Snapshot Serengeti dataset was trained and tested using the Panama dataset.

\begin{table}[h]
\centering
\caption{Architectures used in the experiments}
\label{tab:convnets}
\begin{tabular}{ccc}
\hline
\textbf{Label} & \textbf{Architecture} & \textbf{\# layers} \\
\hline
A         & AlexNet  & 8                                                                  \\
B        & VGG Net               & 16                                                                   \\
C         & GoogLenet             & 22                                                                    \\
D      & ResNet-50             & 50                                                                   \\
E        & ResNet-101            & 101                                                                  \\
F       & ResNet-152            & 152 \\
G         & AlexNet-FT  & 8                                                                
\\
H         & GoogLenet-FT             & 22                                                                  
\\
\hline                                                                
\end{tabular}
\end{table}

Accuracy in the validation set is used as performance metric. Top-1 and Top-5 accuracy are presented as a way to tell how well the model is ranking possible species, and also to explore automatic species classification as a helper when the number of species increases. This metric is a common choice in recognition tasks like the ImageNet recognition challenge~\cite{ILSVRC15}.

\subsection{Implementation Details}

All the dataset images were resized to fit in the ConvNet topologies input: AlexNet  (227x227), VGGNet (224x224), GoogLenet (224x224), and ResNets (224x224). To use ConvNets as feature extractors the last full connected layer was modified to deal with 26 classes instead of 1000 Imagenet challenge classes. Hence, all last full connected layers from all topologies were replaced to deal with $26$ classes.

All used architectures were pre-trained with the ImageNet dataset~\cite{ILSVRC15}. The Fine-tuning process was done running the the back-propagation algorithm with Stochastic gradient descent in the pre-trained topologies. Both the learning rate and step size were reduced in order to deal with overfitting and let the network learn slowly. To fine-tune the models the learning rate of each layer was modified by one layer per test. After the performance is calculated one more layer is allowed to learn and the process is repeated until the performance stops increasing. The process of fine-tuning begins from last to first layer. This is based on  Zeiler et.al's~\cite{zeiler2014visualizing} work, which suggests that the last layers learn more specific class representations. In the pre-trained models the last layers have a specialization in $1000$ ImageNet classes. Ideally, this knowledge will be replaced with $26$ snapshot Serengeti classes through fine-tuning procedure. 

The implementation was done in  the deep learning framework Caffe~\cite{jia2014caffe}, as well as all pre-trained models that were found in the Caffe model Zoo. The Caffe models, the configuration files and the Snapshot Serengeti dataset version used in this paper (including manual segmentation and foreground images) are publicly available for benchmarking at https://sites.google.com/site/udeacameratrap/.

\section{Results}
\label{sec:Results}

\begin{figure}[]
\centering			
\subfigure[]{\includegraphics[width=0.49\textwidth]{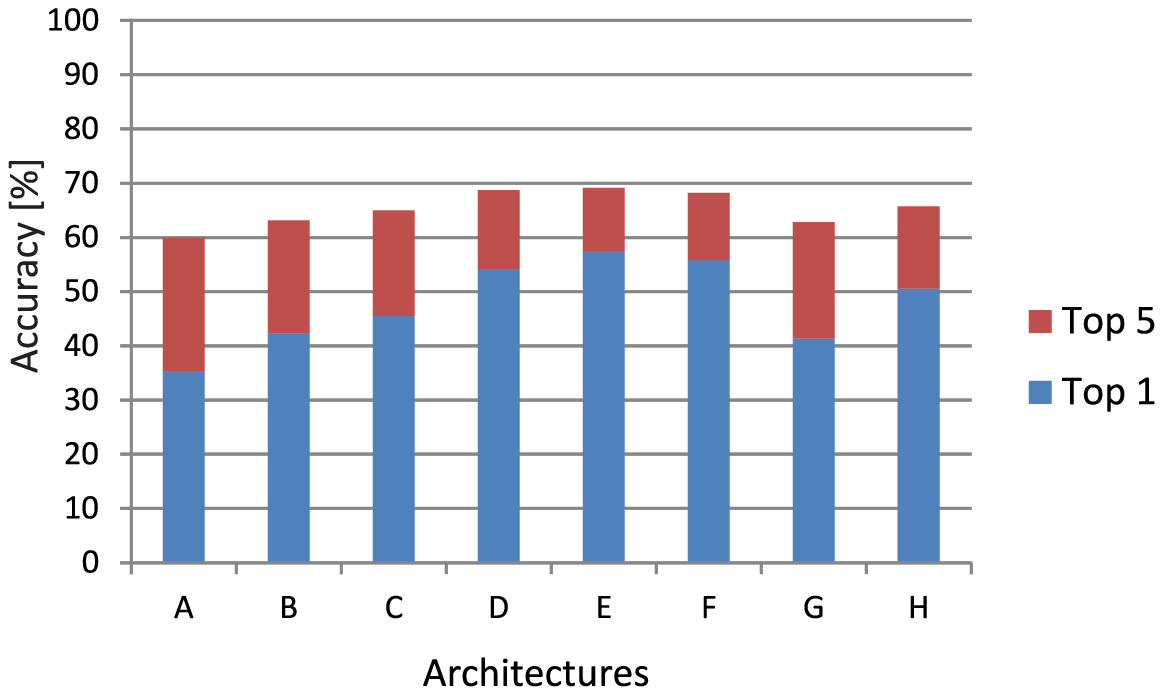}{\label{fig:D1}}}
\subfigure[]{\includegraphics[width=0.49\textwidth]{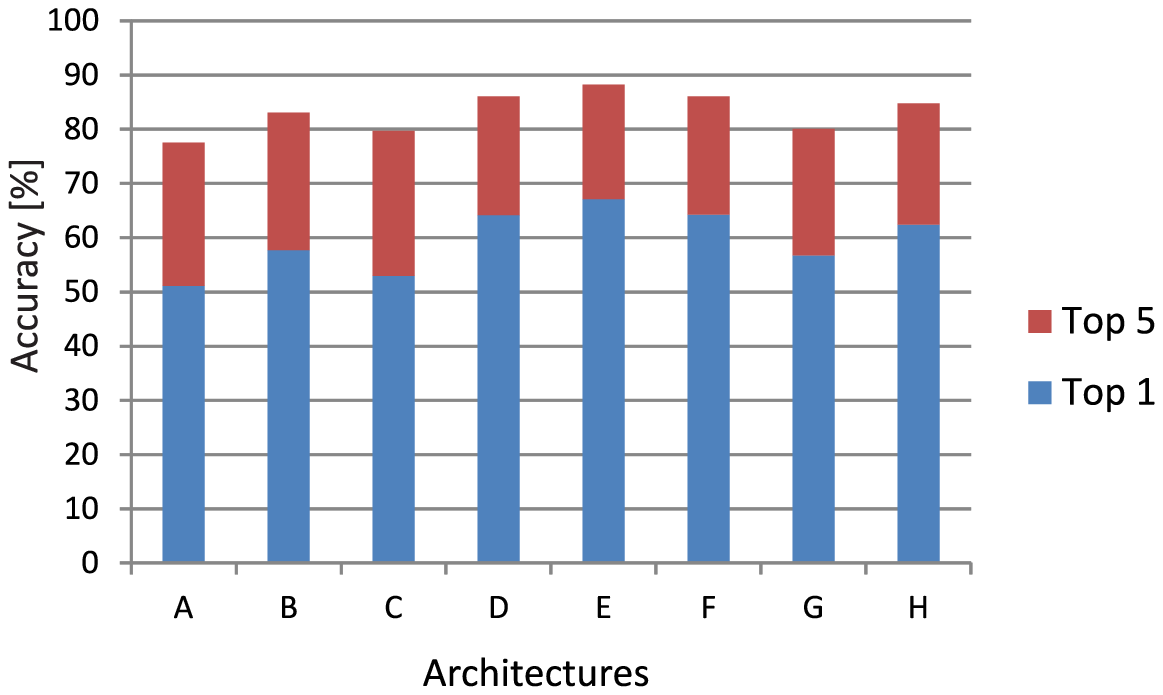}{\label{fig:D2}}}\\
\subfigure[]{\includegraphics[width=0.49\textwidth]{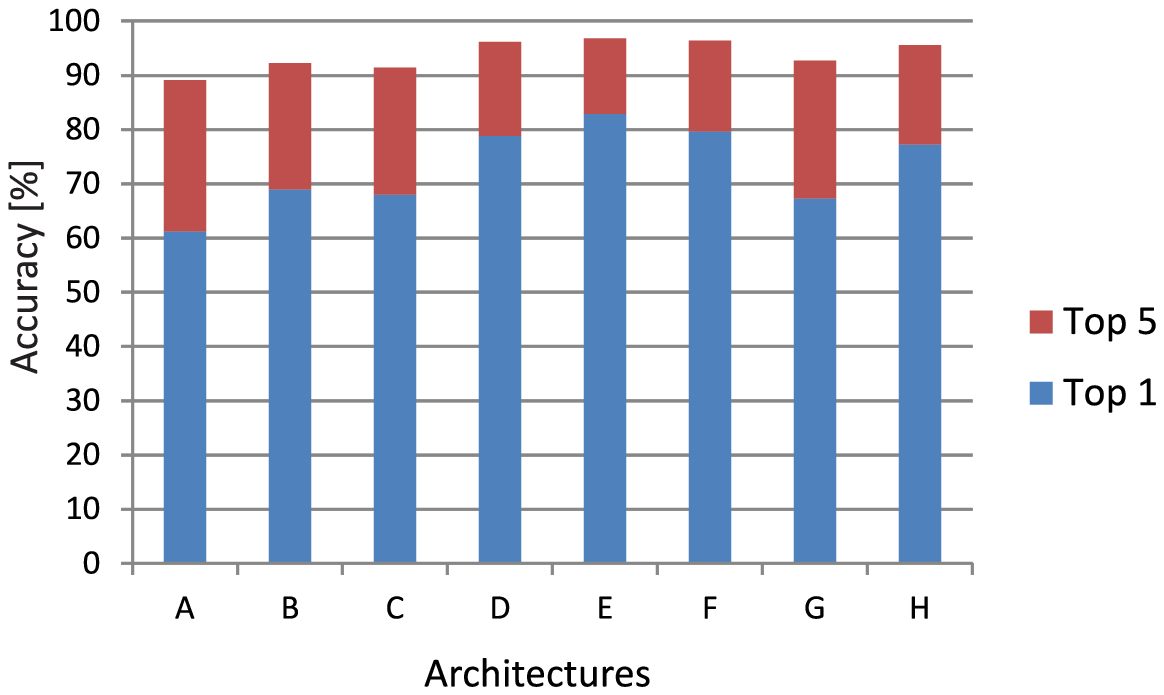}{\label{fig:D3}}}
\subfigure[]{\includegraphics[width=0.49\textwidth]{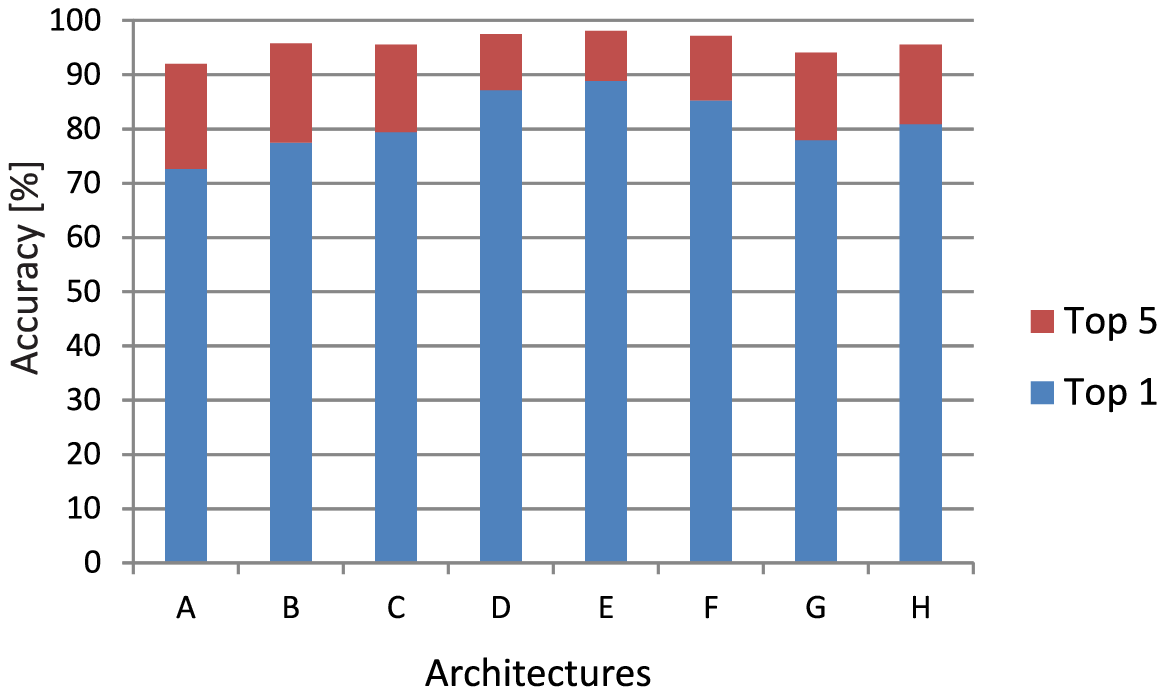}{\label{fig:D4}}}
\caption{Results of the experiments carried out over the Snapshot Serengeti dataset. (a) Using $D1$ dataset. (b) Using $D2$ dataset. (c) Using $D3$ dataset. (d) Using $D4$ dataset.}
\label{fig:results}
\end{figure}

The Fig.~\ref{fig:results} shows the results of experiments using the Snapshot Serengeti dataset. Top-1 and Top-5 results are overlapped for each experiment, for instance for the experiment with dataset $D1$ and architecture A (first bars of Fig.~\ref{fig:D1}), the Top-1  and Top-5 reach an accuracy values of $35.4\%$ and $60.4\% $, respectively.

The experiments with the dataset D1 (see Fig.~\ref{fig:D1}) perform worst. Notice that even the deepest architectures can not deal with this highly unbalanced data. However, the deepest models outperform less deep ones. The results of this experiments arised the question: ``What if biologist collect enough data to build balanced datasets?''

As for experiments with dataset $D2$, the balanced feature is simulated. In this case, the accuracy is higher than for the experiments with dataset $D1$ (see Fig.~\ref{fig:D2}). Similar to the unbalanced case, the better performance was obtained with the deepest topologies. However, the performance still remains under $70\%$ of accuracy. A cause of this results could be that the dataset includes images of both empty frames and animals far away from the camera (in the Snapshot Serengeti dataset this situations are due to the animal's shy nature or camera placement, e.g. the ostrich class). The latter ones become empty frame images when the image is re-sized to fit in the ConvNets. These two issues may be handled by the classifier (as we will discuss later) but only if the number of cases is small. 

Although problems present in the images of the dataset $D2$ depend on time between shots, place of the camera, and sensor parameters, which can be controlled, the animal's behaviour is still an uncontrolled factor. Therefore, this two issues always will be part of the problem and an automatic identification process must consider this issues. 

The experiments with dataset $D3$ show much better performance than the two previous cases, reaching an accuracy of $82.9\%$ and $96.8\%$ for Top-1 and Top-5, respectively (see Fig.~\ref{fig:D3}). This results confirm that the empty frames, which were removed, influence the classification ability of the model. Again, the better results were obtained using the deepest architectures.

The best results were obtained using the dataset $D4$ and the E architecture (Top-1 = $88.9\%$ and Top-5 = $98.1\%$, see Fig.~\ref{fig:D4}). As for the previous cases, the deepest models produced the better results. This is evidence of the ability of the ConvNets arquitectures to deal with the species classification problem, even in cases where the evaluated images contain only parts of the animal body. Although the accuracy values with the dataset $D4$ were the highest, the difference with respect to the ones with the dataset $D3$ is not as noticeable as for the first two sets of experiments (using datasets $D1$ and $D2$). This could be an indicator that the segmentation may not be necessary but a deeper model must be tested instead.

Regarding to the fine-tuned versions of ConvNets, architectures A and C outperformed the black boxes networks in all dataset versions. This result shows how fine-tuning specialized the network on the camera-trap classes and suggests that if deeper architectures are fine-tuned, they probably will outperform our best results. Unfortunately, due to our hardware limitations this hypothesis could not be proved.

\begin{table}[h]
\centering
\caption{Species Recognition Performance on Snapshot Serengeti dataset}
\label{tab:label_test2}
\begin{tabular}{lc@{\hskip 0.5in}lc}
\hline
\multicolumn{1}{c}{Specie}          & Accuracy[\%] & \multicolumn{1}{c}{Specie}           &Accuracy[\%] \\
\hline
Baboon          &  98.3       & Jackal           & 92.4      \\
Buffalo         & 83.3      & Kori bustard       & 97.9      \\
Cheetah         & 97.0       & Lion female\&cub  & 83.3      \\
Dik dik         & 75.8       & Lion Male         & 77.0      \\
Eland           & 87.5       & Ostrich           & 94.1      \\
Elephant        & 90.4      & Reedbuck           & 95.0     \\
Giraffe         & 96.6      & Secretary bird     & 95.4      \\
Grant's Gazelle & 65.0       & Spotted Hyena     & 85.4     \\
Guinea fowl     & 99.5      & Thomson's Gazelle & 71.6    \\
Hartebeest      & 97.5      & Topi              & 95.8      \\
Hippopotamus    & 94.1       & Warthog           & 98.0     \\
Human           & 99.1      & Wildebeest        & 93.1    \\
Impala          & 92.0      & Zebra             & 99.5    \\
\hline
\end{tabular}
\end{table}

The Table~\ref{tab:label_test2} shows accuracy results per class using the architecture E trained with dataset $D4$. In most of the classes the accuracy reached high values. The classes with low performance were the ones related to the fine-grained classification problem.

\begin{table}[]
\centering
\caption{Comparison of Species Recognition Performance on camera-trap dataset~\cite{chen2014deep}.}
\label{tab:label_test}
\begin{adjustbox}{max width=\textwidth}
\begin{tabular}{cccccc}
\hline
Method  & Agouti    & Peccary     & Paca       & R-Brocket Deer & W-nosed Coati \\
\hline
ConvNet~\cite{chen2014deep} & 13        & 12.2        & 18.7       & 2.0            & 24.3          \\
Ours    & 51.2      & 69.6        & 45.2       & 13.8           & 63.0          \\
\hline
Method  & Spiny Rat & Ocelot      & R-Squirrel & Opossum        & Bird spec     \\
\hline
ConvNet~\cite{chen2014deep} & 5.0       & 22.4        & 3.8        & 14.7           & 0.1           \\
Ours    & 9.0       & 59.2        & 32.8       & 41.6           & 29.8          \\
\hline
Method  & Tinamou   & W-Tail Deer & Mouflon    & R-Deer         & Roe Deer      \\
\hline
ConvNet~\cite{chen2014deep} & 29.8      & 50.0        & 71.0       & 82.0           & 4.6           \\
Ours    & 59.2      & 63.3        & 87.2       & 60.7           & 50.8          \\
\hline
Method  & Wild Boar & R-Fox       & Euro Hare  & Wood Mouse     & Coiban Agouti \\
\hline
ConvNet~\cite{chen2014deep} & 17.1      & 1.0         & 2.0        & 87.3           & 4.5           \\
Ours    & 44.3      & 6.4         & 6.3        & 86.9           & 53.8         \\
\hline
\end{tabular}
\end{adjustbox}
\end{table}

The Table~\ref{tab:label_test} presents the results of experiment C1. The model used in this experiments was the ResNet-101 (architecture E). In most of the cases, the results were better, except for the Red Deer and Wood Mouse. These results show how a deeper architecture outperforms the less deeper one used in~\cite{chen2014deep}. As was stated before, this proves that the automation of camera-trap species recognition depends not only on having enough data but also on having a powerful learning algorithm.


\section{Discussion} 
\label{sec:discussion}

\begin{figure}[h]
\centering			
\subfigure[]{\includegraphics[width=0.20\textwidth,height=2cm]{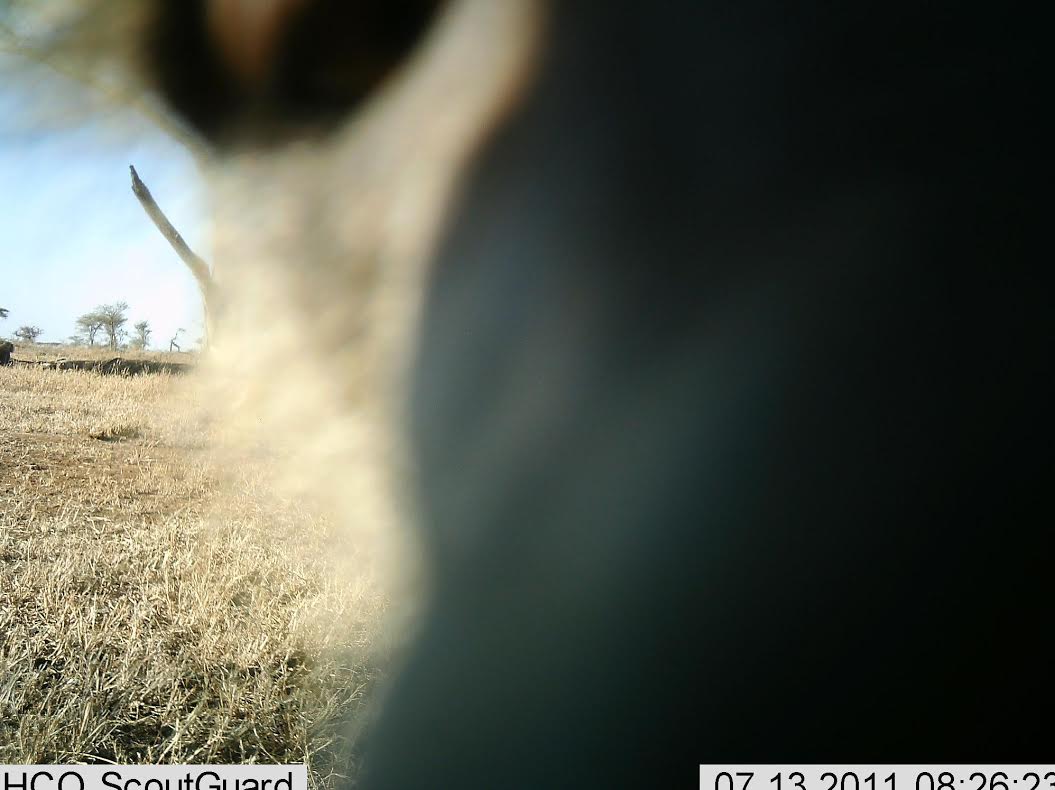}{\label{fig:baboree}}}
\subfigure[]{\includegraphics[width=0.20\textwidth,height=2cm]{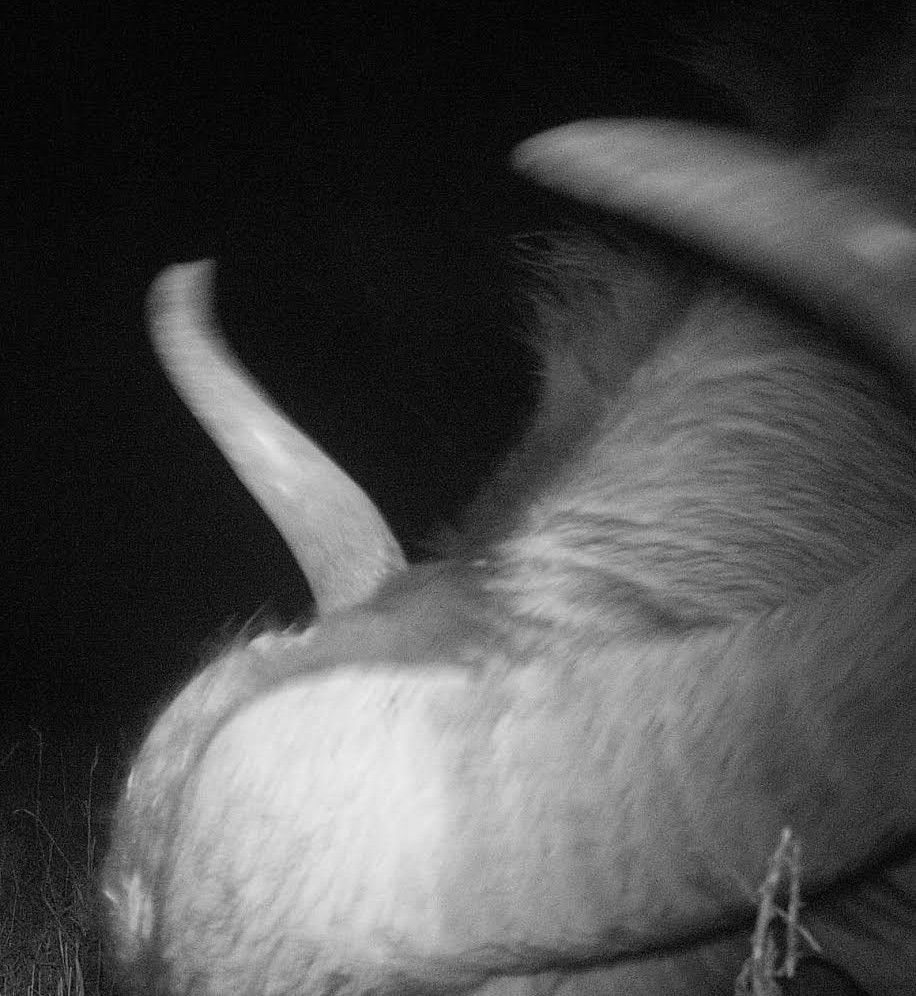}{\label{fig:buffwil}}}
\subfigure[]{\includegraphics[width=0.20\textwidth,height=2cm]{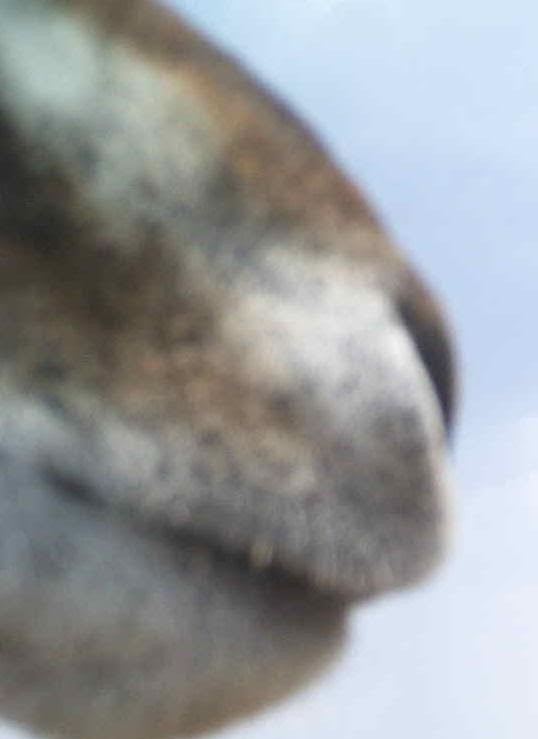}{\label{fig:buffwil2}}}
\subfigure[]{\includegraphics[width=0.20\textwidth,height=2cm]{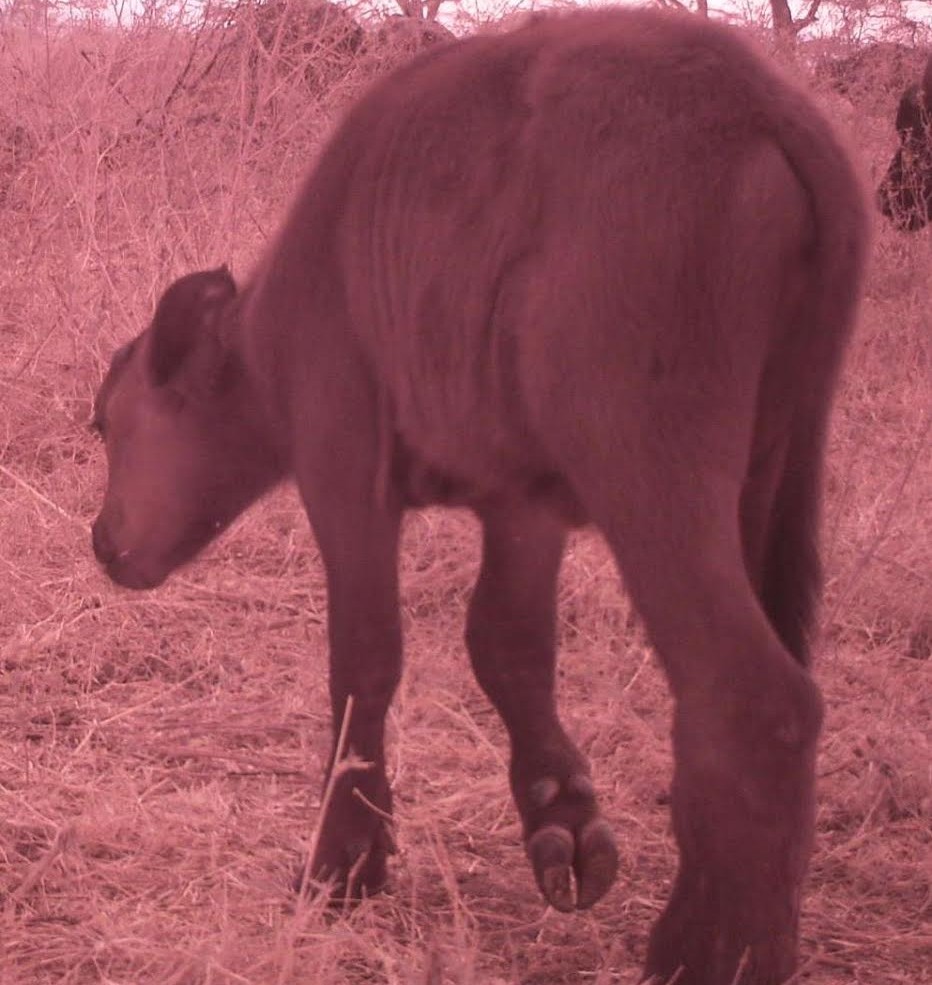}{\label{fig:grantzGir}}}
\subfigure[]{\includegraphics[width=0.20\textwidth,height=2cm]{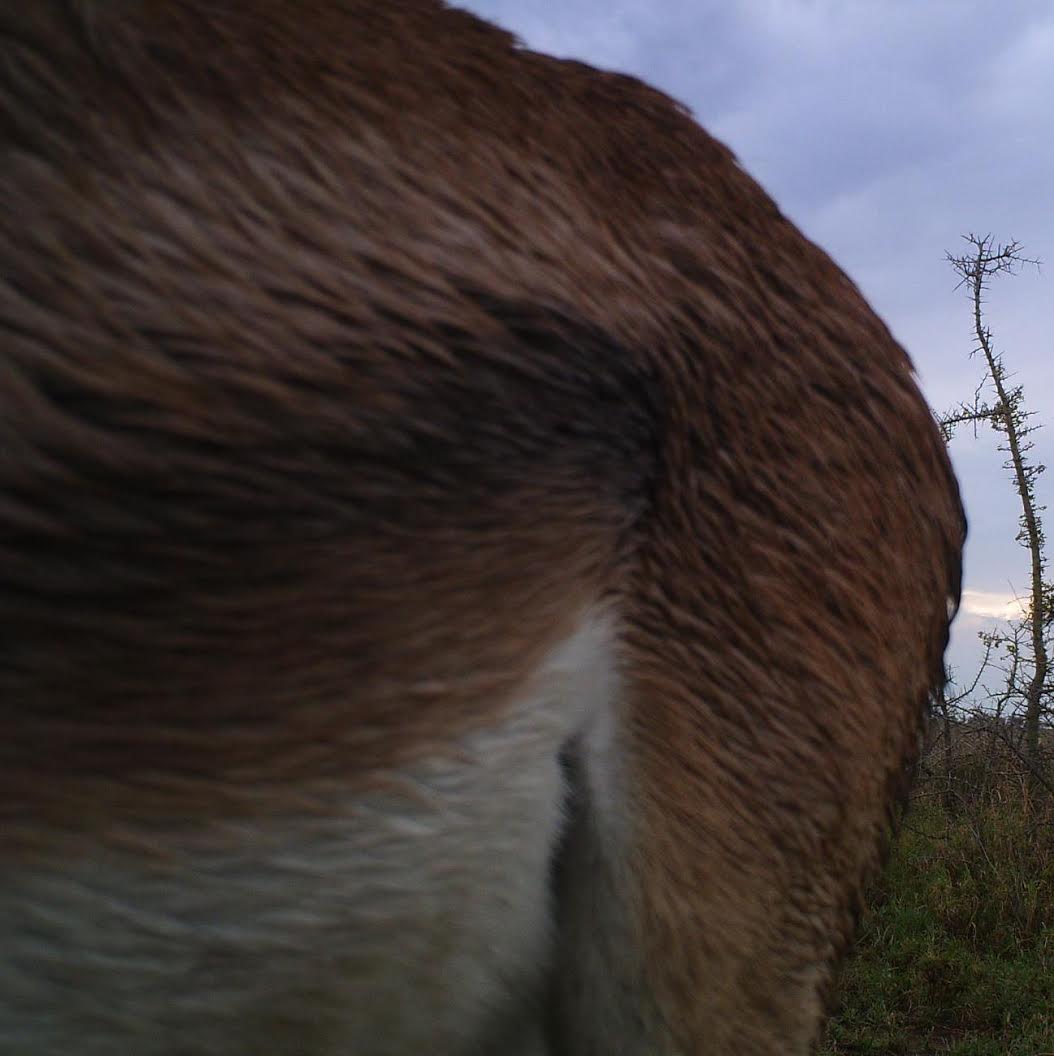}{\label{fig:grantzTho}}}
\subfigure[]{\includegraphics[width=0.20\textwidth,height=2cm]{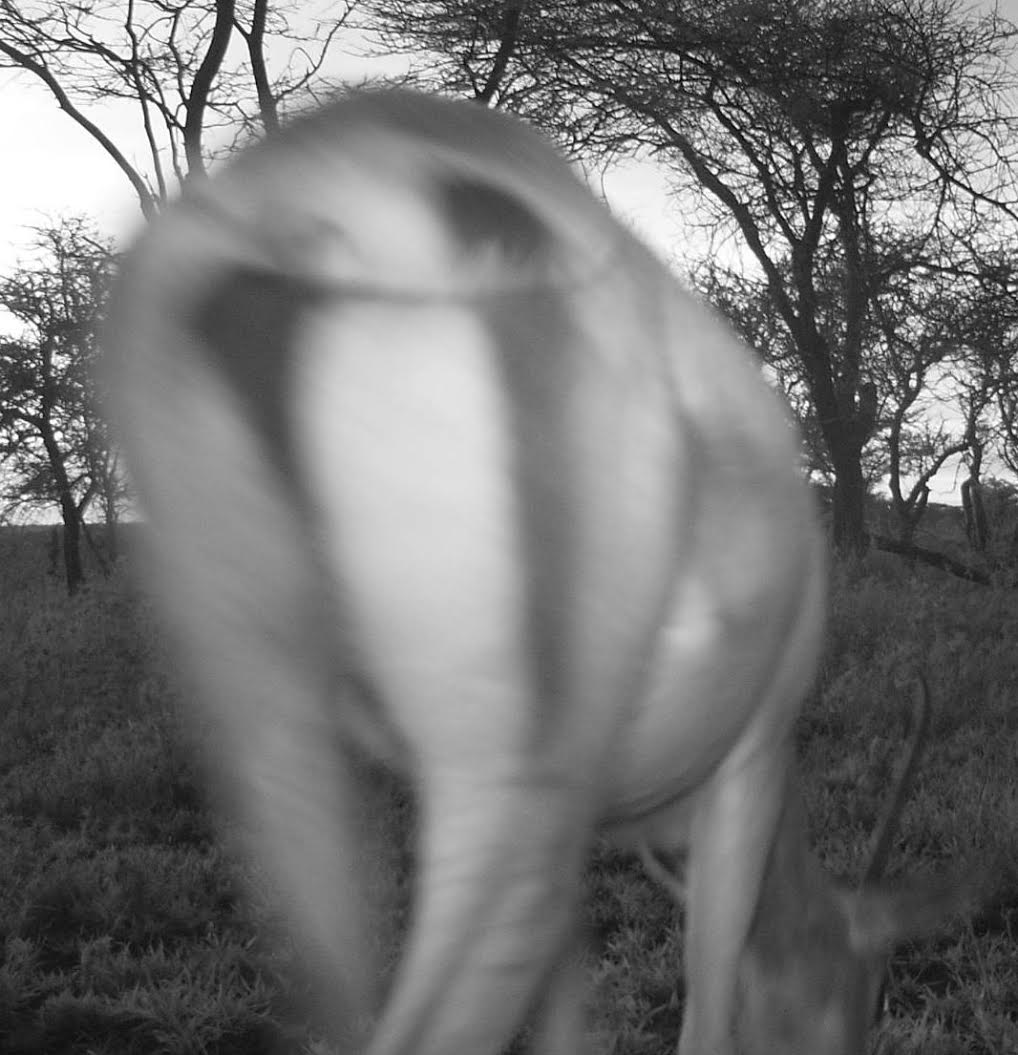}{\label{fig:grantzTho2}}}
\subfigure[]{\includegraphics[width=0.20\textwidth,height=2cm]{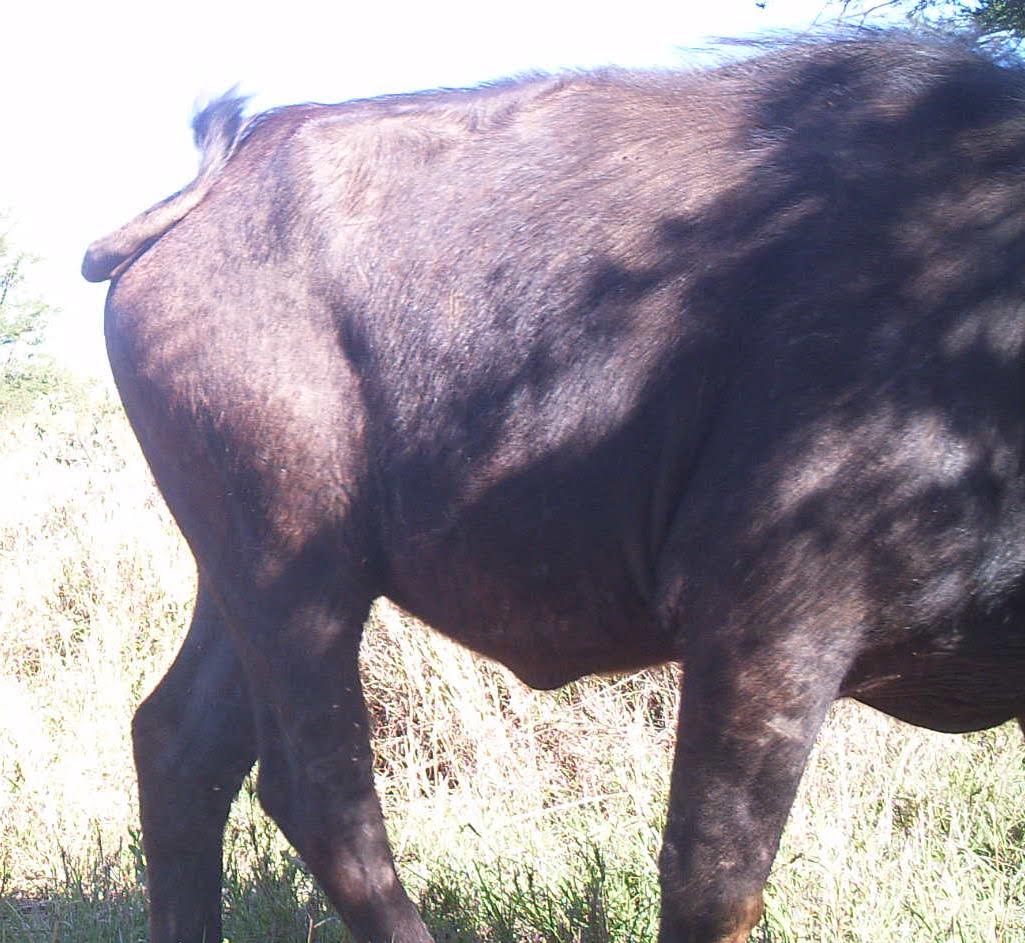}{\label{fig:grantzTho2}}}
\subfigure[]{\includegraphics[width=0.20\textwidth,height=2cm]{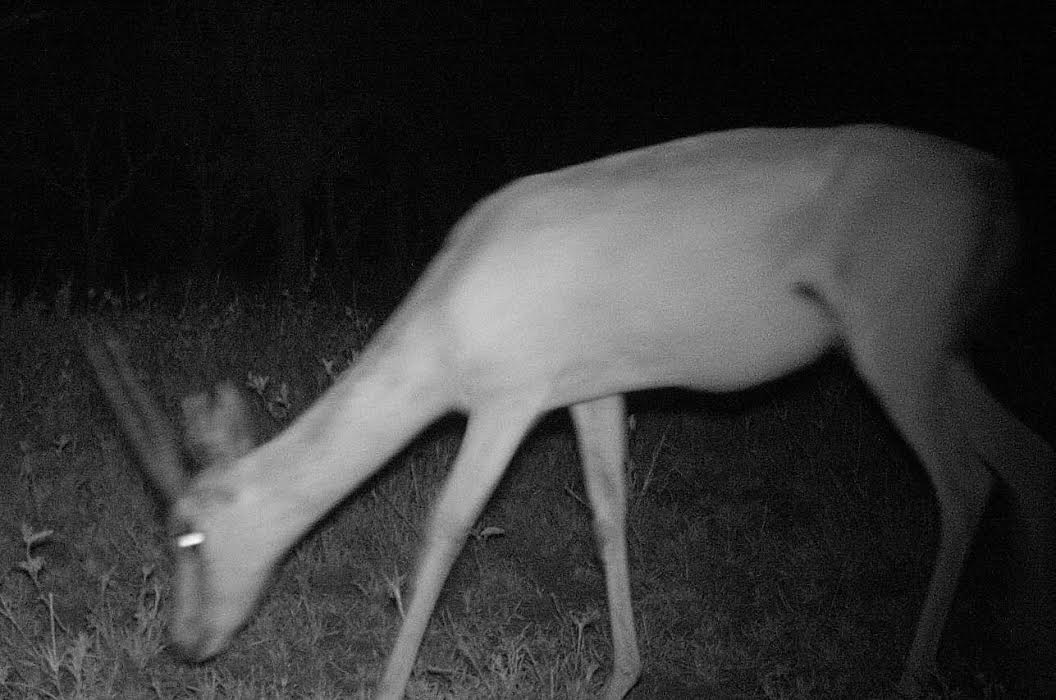}{\label{fig:grantzTho2}}}
\caption{Misclassified images. (a) Baboon classified as impala. (b) Buffalo classified as wildebeest. (c) Grant's Gazelle classified as Giraffe. (d) Buffalo classified as Hyena. (e) Grant's Gazelle classified as Thomson's Gazelle. (f) Grant's Gazelle classified as Impala. (g) Buffalo classified as Wildebeest. (h) Grant's Gazelle classified as Hyena.
}
\label{fig:errors}
\end{figure}

In this work, models (ConvNets) that can classify $1000$ classes with a $19.38\%$ of error (as was reported in~\cite{He2015}) are used. But even more conditioned versions of the dataset could not reach an accuracy higher than $88.9\%$. If our task is a $26$ classes problem, which in some sense is simpler than the $1000$ ImageNet identification challenge, what is the big deal?
Fig.~\ref{fig:errors} shows some classification errors in the evaluation set. From Fig.~\ref{fig:baboree} to Fig.~\ref{fig:grantzGir} exists the same condition presented in Fig.~\ref{fig:impalaLeg} where the models do not have enough information to predict the correct class. This error shows how the ConvNet is specialized in certain parts of animals to recognize the species and its presence or absence influences the classification decision.

In Fig.~\ref{fig:grantzTho} the fine-grained problem previously explained appears. The ConvNet assumes a black line in the body as a feature of Thomson's Gazelle but this is not a deterministic difference between Grant's and Thomson's Gazelle (this black line can also appear in Grant's Gazelle ,perhaps it is rare). In Fig.~\ref{fig:grantzTho2} the fine-grained classification is even harder, due to the fact that this back pattern is similar in Grant's Gazelle and impala species.

The case shown in Fig.~\ref{fig:grantzTho2} exhibits how the ConvNet has to learn more specific features to correctly predict classes in this poses. In an extreme case the ConvNet has to take a low trusted decision if the visual information is not enough. There are a lot of images that are in previously mentioned conditions. Top-5 accuracy in Fig.~\ref{fig:results} shows a high accuracy, which evidences that even if the prediction was wrong the ConvNet is near the correct answer and learned a correct and discriminative patterns for each species.

Fig.~\ref{fig:buffwil2} shows a buffalo's cub. The appearance of the species can be different in young age. The cub of a buffalo does not have the distinctive pattern learned by the model (for buffalo class). This issue arises the necessity of samples from cubs, female, and male individuals of the same species.

In addition, the fact that Fig.~\ref{fig:buffwil2} was classified as hyena is related to two issues. First, the shape of the animal species (buffalo cub as hyena or buffalo as hippopotamus) could resemble. Second,  the image was taken with a damaged camera, mostly all images from hyenas are in gray scale (Hyenas are used to be captured at night). This issue bias the ConvNet to classify all gray scale images  with a ``Hyena'' shape as hyena or ``Hippopotamus'' shape as hippopotamus. To deal with this problem the Hyena class must be provided with more color images.

The Snapshot Serengeti dataset was annotated by citizens and experts. Although the comparison between experts and amateurs gives high accuracy, some mistakes occurred. In~\cite{van2015building} it is discussed how learning algorithms are robust to annotation errors and training data corruption. This assumption is true only if the  training set has sufficient samples. In this work another citizen annotated dataset was used and the results show  robustness to data corruption. Hence, these results confirm  the discussion of~\cite{van2015building} and motivate the use of citizen science. How much impact data corruption has in training, evaluation, and how big the dataset has to be to deal with this annotation errors, are important questions that must be answered.

\section{Conclusions}
\label{sec:Conclusions}

In this paper an automatic species recognition method based on very deep convolutional neural networks is proposed. Extensive experiments using four different versions of the Snapshot Serengeti dataset (that reflects possible automatized scenarios for camera-trap framework) and six state of the art very deep convolutional neural networks, were carried out. Our method achieves $88.9\%$ of accuracy in the Top-1  and $98.1\%$ in the Top-5 in the evaluation set. The results show to which extend learning algorithms are robust to wrong annotated data by citizens. Also multiple problems in the classification task are depicted, such as the intra-class variation due to age or sex appearance variation. The experiments exhibit that it is possible to obtain highly accurate classification of species from camera-trap images, if there is sufficient data and an accurate segmentation algorithm available. The versions of the Snapshot Serengeti dataset, as well as manually segmented images, and deep architecture models are publicly available for benchmarking and make the use of the Snapshot Serengeti dataset for computer vision research easier. In addition, a comparison with a previous approach~\cite{chen2014deep} is drawn. The presented method outperforms the previous results in most of the cases.

In the future, the system will be tested using temporal information. Since camera trapping framework usually offers a burst of images (currently, each image is analysed independently), most of the problems explained in Section~\ref{sec:Methods} could be solved, if the sequence of images is evaluated as a single instance (as biologist do). A new dataset will be produced, which will be splitted in sets corresponding the problems mentioned in Fig.~\ref{fig:preprocessing}. 

Since the classification of camera-trap images requires fine-grained classification, training images using parts of animals will be included into the models. This will allow us to use ImageNet images to increase the training set. 

Finally, one of our short-term goals is to develop a segmentation algorithm that approximates to manual segmentation. 

\bibliographystyle{splncs}
\bibliography{egbib}

\begin{thebibliography}{10}

\bibitem{shiras1906photographing}
Shiras, G.:
\newblock Photographing wild game with flashlight and camera.
\newblock National Geographic Society (1906)

\bibitem{o2010camera}
O'Connell, A.F., Nichols, J.D., Karanth, K.U.:
\newblock Camera traps in animal ecology: methods and analyses.
\newblock Springer Science \& Business Media (2010)

\bibitem{fegraus2011data}
Fegraus, E.H., Lin, K., Ahumada, J.A., Baru, C., Chandra, S., Youn, C.:
\newblock Data acquisition and management software for camera trap data: A case
  study from the team network.
\newblock Ecological Informatics \textbf{6}(6) (2011)  345--353

\bibitem{yu2013automated}
Yu, X., Wang, J., Kays, R., Jansen, P.A., Wang, T., Huang, T.:
\newblock Automated identification of animal species in camera trap images.
\newblock EURASIP Journal on Image and Video Processing \textbf{2013}(1) (2013)
   1--10

\bibitem{chen2014deep}
Chen, G., Han, T.X., He, Z., Kays, R., Forrester, T.:
\newblock Deep convolutional neural network based species recognition for wild
  animal monitoring.
\newblock In: 2014 IEEE International Conference on Image Processing (ICIP),
  IEEE (2014)  858--862

\bibitem{swanson2015snapshot}
Swanson, A., Kosmala, M., Lintott, C., Simpson, R., Smith, A., Packer, C.:
\newblock Snapshot serengeti, high-frequency annotated camera trap images of 40
  mammalian species in an african savanna.
\newblock Scientific data \textbf{2} (2015)

\bibitem{krizhevsky2012imagenet}
Krizhevsky, A., Sutskever, I., Hinton, G.E.:
\newblock Imagenet classification with deep convolutional neural networks.
\newblock In: Advances in neural information processing systems. (2012)
  1097--1105

\bibitem{simonyan2014very}
Simonyan, K., Zisserman, A.:
\newblock Very deep convolutional networks for large-scale image recognition.
\newblock arXiv preprint arXiv:1409.1556 (2014)

\bibitem{szegedy2015going}
Szegedy, C., Liu, W., Jia, Y., Sermanet, P., Reed, S., Anguelov, D., Erhan, D.,
  Vanhoucke, V., Rabinovich, A.:
\newblock Going deeper with convolutions.
\newblock In: Proceedings of the IEEE Conference on Computer Vision and Pattern
  Recognition. (2015)  1--9

\bibitem{He2015}
He, K., Zhang, X., Ren, S., Sun, J.:
\newblock Deep residual learning for image recognition.
\newblock arXiv preprint arXiv:1512.03385 (2015)

\bibitem{lecun1998gradient}
LeCun, Y., Bottou, L., Bengio, Y., Haffner, P.:
\newblock Gradient-based learning applied to document recognition.
\newblock Proceedings of the IEEE \textbf{86}(11) (1998)  2278--2324

\bibitem{glorot2010understanding}
Glorot, X., Bengio, Y.:
\newblock Understanding the difficulty of training deep feedforward neural
  networks.
\newblock In: International conference on artificial intelligence and
  statistics. (2010)  249--256

\bibitem{razavian2014cnn}
Razavian, A., Azizpour, H., Sullivan, J., Carlsson, S.:
\newblock Cnn features off-the-shelf: an astounding baseline for recognition.
\newblock In: Proceedings of the IEEE Conference on Computer Vision and Pattern
  Recognition Workshops. (2014)  806--813

\bibitem{pan2010survey}
Pan, S.J., Yang, Q.:
\newblock A survey on transfer learning.
\newblock Knowledge and Data Engineering, IEEE Transactions on \textbf{22}(10)
  (2010)  1345--1359

\bibitem{ren2013ensemble}
Ren, X., Han, T., He, Z.:
\newblock Ensemble video object cut in highly dynamic scenes.
\newblock In: Proceedings of the IEEE Conference on Computer Vision and Pattern
  Recognition. (2013)  1947--1954

\bibitem{ILSVRC15}
Russakovsky, O., Deng, J., Su, H., Krause, J., Satheesh, S., Ma, S., Huang, Z.,
  Karpathy, A., Khosla, A., Bernstein, M., Berg, A.C., Fei-Fei, L.:
\newblock {ImageNet Large Scale Visual Recognition Challenge}.
\newblock International Journal of Computer Vision (IJCV) \textbf{115}(3)
  (2015)  211--252

\bibitem{zeiler2014visualizing}
Zeiler, M.D., Fergus, R.:
\newblock Visualizing and understanding convolutional networks.
\newblock In: Computer vision--ECCV 2014.
\newblock Springer (2014)  818--833

\bibitem{jia2014caffe}
Jia, Y., Shelhamer, E., Donahue, J., Karayev, S., Long, J., Girshick, R.,
  Guadarrama, S., Darrell, T.:
\newblock Caffe: Convolutional architecture for fast feature embedding.
\newblock arXiv preprint arXiv:1408.5093 (2014)

\bibitem{van2015building}
Van~Horn, G., Branson, S., Farrell, R., Haber, S., Barry, J., Ipeirotis, P.,
  Perona, P., Belongie, S.:
\newblock Building a bird recognition app and large scale dataset with citizen
  scientists: The fine print in fine-grained dataset collection.
\newblock In: Proceedings of the IEEE Conference on Computer Vision and Pattern
  Recognition. (2015)  595--604

\end{thebibliography}
\end{document}